%% file: main.tex
\documentclass[format=acmsmall,review=false, screen=true,nonacm]{acmart}

\usepackage{booktabs}
\usepackage[ruled]{algorithm2e}
\graphicspath{{Figures}}
\DeclareGraphicsExtensions{.pdf,.png}
\usepackage{bigstrut}
\usepackage{multirow}
\usepackage{multicol}

\newcommand{\p}[1]{\smallskip \noindent \textbf{{#1}.}}
\newcommand{\eq}[1]{Equation~(\ref{eq:#1})}
\newcommand{\fig}[1]{Figure~\ref{fig:#1}}

\usepackage{graphicx}
\graphicspath{{Figures}}
\DeclareGraphicsExtensions{.pdf,.png}

\usepackage{csquotes}
\usepackage{balance}
\usepackage{amsmath}
\usepackage{bm}

\setcopyright{none}

\begin{document}

\title{SARI: Shared Autonomy across Repeated Interaction}


\author{Ananth Jonnavittula}
\orcid{0000-0002-0711-2051}
\email{ananth@vt.edu}

\author{Shaunak A. Mehta}
\orcid{0000-0002-3160-6879}
\email{mehtashaunak@vt.edu}

\author{\\Dylan P. Losey}
\orcid{0000-0002-8787-5293}
\email{losey@vt.edu}

\affiliation{
  \institution{Virginia Tech}
  \department{Department of Mechanical Engineering}
  \city{Blacksburg}
  \state{VA}
  \postcode{24060}
  \country{USA}
  }

\renewcommand\shortauthors{A. Jonnavittula et al.}

\begin{abstract}

Assistive robot arms try to help their users perform everyday tasks. One way robots can provide this assistance is \textit{shared autonomy}. Within shared autonomy, both the human and robot maintain control over the robot’s motion: as the robot becomes confident it understands what the human wants, it intervenes to automate the task. But how does the robot know these tasks in the first place? State-of-the-art approaches to shared autonomy often rely on prior knowledge. For instance, the robot may need to know the human's potential goals beforehand. During long-term interaction these methods will inevitably break down --- sooner or later the human will attempt to perform a task that the robot does not expect. Accordingly, in this paper we formulate an alternate approach to shared autonomy that learns assistance from scratch. Our insight is that operators \textit{repeat} important tasks on a daily basis (e.g., opening the fridge, making coffee). Instead of relying on prior knowledge, we therefore take advantage of these repeated interactions to learn assistive policies. We introduce SARI, an algorithm that \textit{recognizes} the human’s task, \textit{replicates} similar demonstrations, and \textit{returns} control when unsure. We then combine learning with control to demonstrate that the error of our approach is uniformly ultimately bounded. We perform simulations to support this error bound, compare our approach to imitation learning baselines, and explore its capacity to assist for an increasing number of tasks. Finally, we conduct three user studies with industry-standard methods and shared autonomy baselines, including a pilot test with a disabled user. Our results indicate that learning shared autonomy across repeated interactions matches existing approaches for known tasks and outperforms baselines on new tasks. See videos of our user studies here: \url{https://youtu.be/3vE4omSvLvc}

\end{abstract}




\maketitle

\input{intro}

\input{related}

\input{problem}
\input{method}
\input{theory}
\input{simulations}
\input{userstudy}
\input{conclusion}
\input{appendix}


\balance
\bibliographystyle{ACM-Reference-Format}
\bibliography{bibtex}

\end{document}

%% file: intro.tex
\section{Introduction}

\begin{figure}[t!]
	\begin{center}
		\includegraphics[width=0.5\columnwidth]{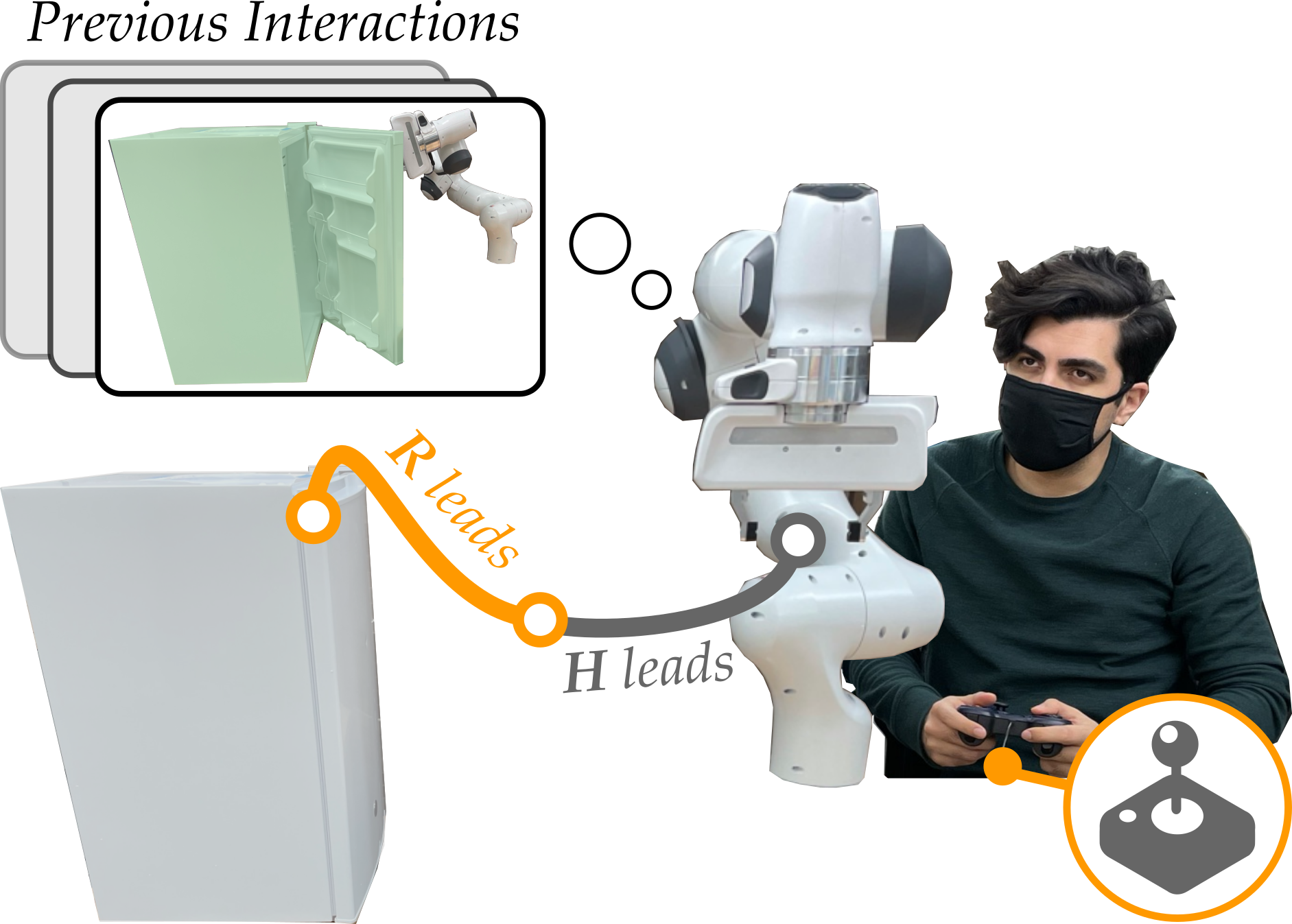}

		\caption{User teleoperating an assistive robot arm to open their fridge door. The robot does not have any prior knowledge about this task; however, the human and robot have completed similar tasks many times before. Instead of making the human guide the robot through every step of this task, we hypothesize that robot arms can learn to assist humans and share autonomy by exploiting the \textit{repeated} nature of everyday tasks.}
		\label{fig:front}
	\end{center}

\end{figure}

Imagine teleoperating a wheelchair-mounted robot arm to open your refrigerator door (see \fig{front}). This robot has never interacted with your fridge before: accordingly, for the first few times you open the fridge, you must carefully guide the robot throughout the entire process of reaching, grabbing, and pulling the door. But after you've interacted with this robot for several weeks --- and opened your fridge \textit{many} times --- an intelligent robot should learn to \textit{assist} you. The next time you start teleoperating the arm towards your fridge, the robot should recognize what you want and partially automate the process of opening the door.

The robot's assistance in this working example is an instance of \textit{shared autonomy}. Shared autonomy for assistive robot arms blends the user's inputs with autonomous actions so that both the human and the robot contribute to the robot's overall motion. When surveyed, disabled adults who operate assistive robots preferred shared autonomy over either fully autonomous or fully human-guided systems \cite{bhattacharjee2020more, gopinath2016human}. In practice, however, today's shared autonomy approaches rely on \textit{prior information} about the human's tasks. Methods such as \cite{dragan2013policy, jain2019probabilistic, javdani2018shared, brooks2019balanced, newman2018harmonic, nikolaidis2017human, aronson2018eye, admoni2016predicting, jonnavittula2021know} require a pre-defined list of goals the human might want to reach: the robot infers the human's most likely goal from these discrete options, and autonomously moves towards that goal. Other approaches need demonstrations \cite{losey2022learning, jeon2020shared, mehta2021learning}, feedback \cite{reddy2018shared}, or constraints \cite{schaff2020residual, bragg2020fake, broad2020data, du2020ave} that specify which tasks the human might want to perform: the robot then maps the human's inputs to task relevant motions, and overrides or corrects inputs that do not align with the robot's anticipated tasks.

These existing approaches work well when the user wants to perform a task that the robot knows \textit{a priori}. But what happens when the human inevitably wants to complete some new or unexpected task? Going back to our working example, the robot has no prior information about opening the refrigerator. When the user teleoperates the robot towards the fridge door, today's assistive arms assume that the human has made a mistake: instead of helping for the fridge task, shared autonomy guides the robot towards one of its known tasks. Even worse, the robot remains confused --- and provides incorrect assistance --- no matter how many times the operator tries to repeat the process of opening the fridge \cite{zurek2021situational}.

For assistive arms to be practical across long-term interaction, these robots must be capable of \textit{learning} a spectrum of new tasks. This would be extremely challenging if every task was a unique one-off that the robot had never seen before. But our insight is that, over the many weeks, months, and years a human operator works with their assistive robot:
\begin{center}
\textit{Humans constantly} repeat \textit{tasks that are important in their everyday life.}
\end{center}
We emphasize that these repetitions are never exactly the same. Each time the assistive robot opens the refrigerator it may have a different start position or follow a different trajectory. Hence, we cannot simply record and playback the motions that the human has shown --- instead, we need to generalize assistance across similar tasks. Applying our insight enables assistive robot arms to learn to share autonomy by exploiting the \textit{repeated interactions} inherent within assistive applications. Here the robot remembers how the user controlled the arm to open the fridge in the past, recognizes that the user is providing similar inputs during the current interaction, and assists by autonomously mimicking the behavior that user previously demonstrated. Across repeated human-robot interactions these assistive arms should learn to share autonomy for tasks that include not only \textit{discrete goals} (e.g., reaching a cup) but also \textit{continuous skills} (e.g., opening a door).

In this paper we propose, analyze, and evaluate our algorithm \textbf{SARI}: \textbf{S}hared \textbf{A}utonomy across \textbf{R}epeated \textbf{I}nteraction. Overall, we make the following contributions\footnote{Parts of this work have been previously published in the IEEE/RSJ International Conference on Intelligent Robots and Systems \cite{jonnavittula2021learning}. Novel contributions include the combination of learning and control (Section~\ref{sec:theory} and Appendix), experiments on SARI capacity in Section~\ref{sec:sim_capacity}, and two user studies in Section~\ref{sec:user}. This added material provides formal guarantees about the performance of our approach, offers a new understanding of how our approach works in practice, and compares our method's performance to state-of-the-art baselines with non-disabled users and one disabled adult.}:

\p{Capturing Latent Intent} We formalize the problem of sharing autonomy across repeated human-robot interaction. During each interaction the human has in mind some desired task: we introduce an end-to-end imitation learning algorithm that learns to recognize the human's current intent and provide autonomous assistance without any pre-defined tasks or prior information.

\p{Returning Control when Uncertain} Our approach should assist during previously seen tasks without overriding the human whenever they try to perform a new task. We introduce a discriminator to measure the confidence of our learned assistance so that the robot automatically returns control to the human when it is unsure about what the human really wants.

\p{Analyzing Stability} We combine learning with control theory to bound the error between the robot's actual state and the human's desired state. We theoretically demonstrate that --- even in the worst case --- our approach is uniformly ultimately bounded with respect to some radius about the human's goal. We derive this radius as a function of the variance in the human's input commands and the similarity between previously learned task(s) and the human's current task.

\p{Comparing to Baselines} We perform experiments with simulated human operators and real robot arms to demonstrate how each component of our algorithm contributes to its performance. Within these controlled experiments we compare our approach to state-of-the-art imitation learning baselines, and test our method's capacity to learn assistance for an increasing number of tasks.

\p{Conducting User Studies} We assess our resulting algorithm in three separate user studies. The first two studies were performed with non-disabled users, and the final study is a pilot with a disabled user who regularly operates wheelchair-mounted robot arms. In these studies we compare our approach to shared autonomy baselines for tasks that involve discrete goals and continuous skills. Viewed together, our results suggest that SARI enables the robot to learn to assist for known and new tasks, leading to higher objective and subjective performance.

%% file: related.tex
\section{Related Work}

Our approach learns to share autonomy across repeated human-robot interaction without predefined tasks or offline demonstrations. Our work is motivated by assistive applications where disabled users teleoperate robot arms on a daily basis. Instead of forcing the user to repeatedly guide the robot throughout every step of the motion, we learn to recognize the human's task, imitate their previous interactions, and arbitrate control between the human and robot.

\p{Application -- Assistive Robot Arms} Over $13\%$ of American adults living with physical disabilities have difficulty with at least one activity of daily living (ADL) \cite{taylor2018americans}. Assistive robots --- such as wheelchair-mounted robot arms \cite{argall2018autonomy, brose2010role, jaco} --- have the potential to help users perform these everyday tasks without relying on caregivers. Recent work on assistive arms has focused on automating ADLs such as eating dinner \cite{feng2019robot, jeon2020shared, park2020active, belkhale2022balancing}, getting dressed \cite{erickson2020assistive, clegg2020learning, puthuveetil2022bodies}, and manipulating household objects \cite{choi2009list}. For sufficiently simple daily tasks the disabled adult may not require any assistance from the robot. However, our research takes inspiration from the fact that users need assistance when performing \textit{complex} tasks  that are \textit{repeated} on a daily basis (e.g., opening a door). It is mentally tedious and physically burdensome for the human to precisely teleoperate the robot throughout each step of these everyday tasks \cite{herlant2016assistive}. When surveyed, adults who operate assistive arms indicated that they prefer to \textit{share autonomy} with robots \cite{bhattacharjee2020more, gopinath2016human}.

\begin{figure}[t]
	\begin{center}
		\includegraphics[width=0.7\columnwidth]{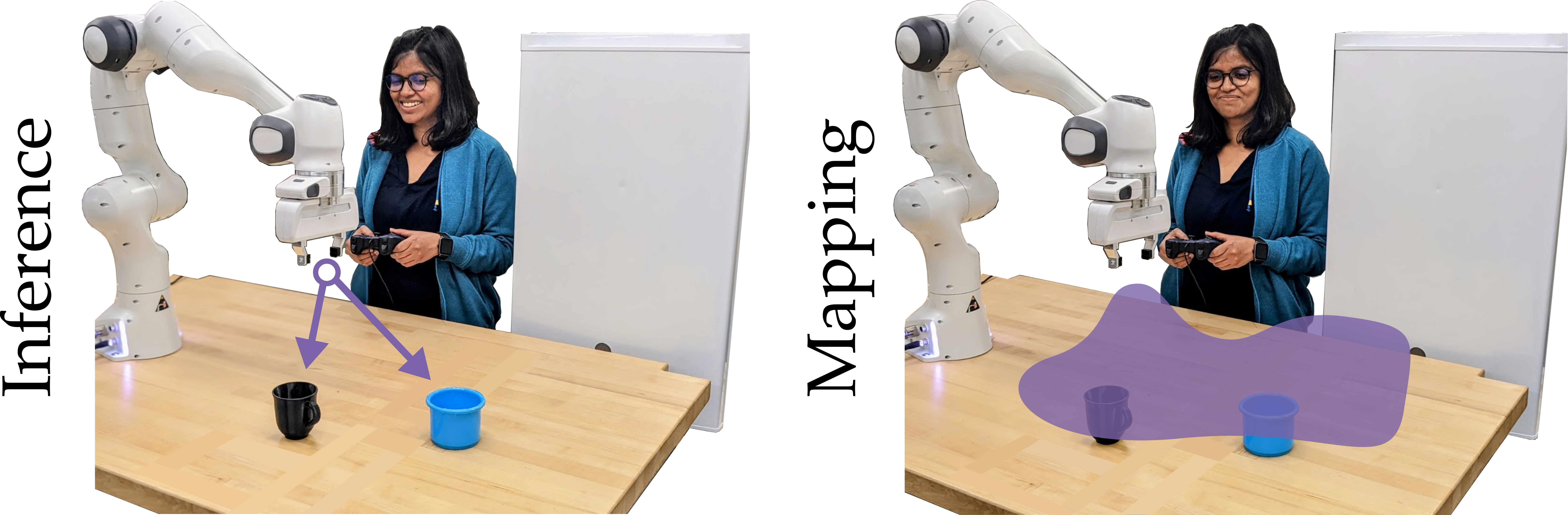}

		\caption{We separate prior work on shared autonomy for assistive robot arms into two groups. (Left) Some methods are given a discrete set of possible goals, and infer the human's goal from these discrete options. (Right) Other methods learn to map the human's joystick inputs to constrained, task-relevant motions. Although both shared autonomy algorithms help this human reach for the cups, neither can assist the human for a new, unexpected task (like opening the fridge).}
		\label{fig:related}
	\end{center}

\end{figure}

\p{Shared Autonomy} In shared autonomy both the human and robot arbitrate control over the robot's motion. We separate related research on shared autonomy into two classes of algorithms (see \fig{related}). First, there are approaches which \textit{infer} the human's goal and then partially automate the arm's motion towards that goal. Second, there are methods which \textit{map} the human's inputs to constrained and task-relevant actions.

Within inference works the robot is given a discrete set of possible goals the human may want to reach \textit{a priori} \cite{dragan2013policy, jain2019probabilistic, javdani2018shared, brooks2019balanced, newman2018harmonic, nikolaidis2017human, aronson2018eye, admoni2016predicting, jonnavittula2022communicating, fuchs2021gaze, mehta2023riso}. Based on the human's teleoperation inputs so far the robot determines which goal(s) are likely, and takes assistive actions to move towards the inferred goal(s). For example, Javdani \textit{et al.} formulate this as a partially observable Markov decision process where the human's goal is the latent state and the human's teleoperation inputs are observations about that latent goal \cite{javdani2018shared}. We emphasize that this class of algorithms \textit{requires prior knowledge} about the human's tasks (i.e., the location of all the goals the human may want to reach). In the long-term these priors will inevitably fall short: sooner or later the human will reach for a goal that the robot did not expect, and the robot will be unable to provide assistance.

Other works on shared autonomy map the human's joystick inputs to motions that are relevant for the current task \cite{bragg2020fake, broad2020data, reddy2018shared, schaff2020residual, losey2022learning, jeon2020shared, mehta2021learning, du2020ave}. When the human provides a suboptimal input (e.g., an input that moves the robot away from its goal) the robot overrides and corrects this human's action. For instance, Reddy \textit{et al.} \cite{reddy2018shared} learn a reward function from the human and then constrain the robot to take actions with high long-term rewards, while Broad \textit{et al.} prevent the human from taking actions that deviate from the robot's expectations. Similarly, Losey \textit{et al.} \cite{losey2022learning} map the human's joystick inputs to latent, task-relevant robot actions that are learned from offline task demonstrations. Overall, this class of algorithms makes sense when the user wants to perform task(s) that the robot has learned to assist. But if the human attempts to perform a new or unexpected task, then these \textit{constraints become counter-productive}: the robot mistakenly overrides the human and may force them to perform the wrong task.

Beyond these two classes of algorithms we highlight recent shared autonomy work that \textit{learns new tasks} during interaction \cite{zurek2021situational, qiao2021learning}. Here the robot starts with a discrete set of options and tries to infer the human's current task. If the human's inputs do not match any of these known tasks, shared autonomy stops: the robot returns full control to the human and the human demonstrates their new task to the robot. This task is then added to the discrete library of options and shared autonomy restarts at the next interaction. Like \cite{zurek2021situational} and \cite{qiao2021learning} our approach continually learns to assist for new tasks. However, we do not \textit{separate} our approach into distinct phases for sharing autonomy or learning tasks. Instead, our robot learns to assist the human each time the user interacts with the robot, regardless of whether they are completing a previously seen task or performing a new skill.

\p{Interactive Imitation Learning} Our technical approach builds on interactive and safe imitation learning \cite{ross2011reduction, osa2018algorithmic}. Specifically, we draw connections between shared autonomy and imitation learning techniques where the robot and human periodically switch control. Within these settings the robot attempts to perform the task autonomously; but at times where the human notices that the robot is making a mistake \cite{spencer2022expert, mandlekar2020human, jauhri2021interactive, kelly2019hg}, or in states where the robot is uncertain about what it should do \cite{zhang2017query, menda2017dropoutdagger, hoque2022thriftydagger, mehta2022unified}, the human takes over and guides the robot. Across repeated interactions the robot adds these human corrections to its training data and learns to imitate the human's behavior. Here we leverage a similar approach to learn to share autonomy. More specifically, our technical approach integrates prior work on both interactive imitation learning and representation learning \cite{jonschkowski2015learning, lynch2020learning, losey2022learning, he2022learning}. We develop representation learning to identify the space of possible tasks, and then incorporate imitation learning to mimic how the human previously performed these tasks and provide autonomous assistance.

%% file: problem.tex
\section{Formalizing Shared Autonomy across Repeated Interaction}

Let us return to our motivating example from \fig{front} where the user is teleoperating their assistive robot arm. Each time the human interacts with the robot, they have in mind a \textit{task} they want the robot to perform: some of these tasks are new (e.g., moving a coffee cup), while others the robot may have seen before (e.g., opening the fridge). We represent the human's current task as $z \in \mathcal{Z}$, so that during interaction $i$, the human wants to complete task $z^i$. Within this paper tasks include both discrete goals and continuous skills: i.e., a task $z$ could be reaching the cup or opening a drawer. We test both types of tasks in our experiments. The assistive robot's goal is to help the human complete their current task. However, the robot does not know (a) \textit{which} task the human currently has in mind or (b) \textit{how} to correctly perform that task.

\p{Dynamics} The robot is in state $s \in \mathbb{R}^d$ and takes action $a\in \mathbb{R}^d$. Within our experiments, $s$ is the robot's joint position, $a$ is the robot's joint velocity, and the robot has dynamics:
\begin{equation} \label{eq:P1}
    s^{t+1} = s^t + \Delta t \cdot a^t
\end{equation}
The human uses a joystick to tell the robot what action to take. Let $a_{\mathcal{H}}$ be the human's \textit{commanded action} --- i.e., the joint velocity corresponding to the human's joystick input\footnote{Although we use joysticks for explanation here, our algorithmic framework is not tied to any teleoperation interface. Users could alternatively control the robot with sip-and-puff devices \cite{jain2019probabilistic}, body-machine interfaces \cite{jain2015assistive}, or brain-computer interfaces \cite{muelling2015autonomy}. All of these interfaces output a commanded action $a_{\mathcal{H}}$. To show our method's ability to generalize, we test using both a joystick and a web-based GUI in our user studies.}. The robot assists the human with an \textit{autonomous action} $a_{\mathcal{R}}$, so that the overall action $a$ is a linear blend of the human's joystick input and the robot's assistive guidance \cite{jain2019probabilistic,dragan2013policy,newman2018harmonic}:
\begin{equation} \label{eq:P2}
    a = \beta \cdot a_{\mathcal{R}} + (1 - \beta) \cdot a_{\mathcal{H}}
\end{equation}
Here $\beta \in [0, 1]$ \textit{arbitrates} control between human and robot. When $\beta \rightarrow0$, the human always controls the robot, and when $\beta \rightarrow1$, the robot acts autonomously.

\p{Human} So how does the human choose inputs $a_{\mathcal{H}}$? During interaction $i$ we assume the human has in mind a desired task $z^i$. We know that this task guides the human's commanded actions; similar to prior work \cite{javdani2018shared}, we accordingly write the \textit{human's policy} as ${\pi_{\mathcal{H}}(a_{\mathcal{H}} \mid s, z^i)}$. This policy is the gold standard, because if we knew $\pi_\mathcal{H}$ we would know exactly how the human likes to perform each task $z \in \mathcal{Z}$. It's important to recognize that this policy is highly \textit{personalized}. Imagine that the current task is to reach a coffee cup at state $s^*$: one human might prefer to move directly towards the cup with actions $a_{\mathcal{H}} \propto (s^* - s)$, while another user takes a circuitous route to stay farther away from obstacles. Our approach should personalize, and learn the policy that the current user prefers.

\p{Repeated Interaction} In practice the assistive robot cannot directly observe either $z^i$ or $\pi_\mathcal{H}$. Instead, the robot observes the states that it visits and the commands that the human provides. Let $\tau = \{(s^1, a_{\mathcal{H}}^1), \ldots, (s^T, a_{\mathcal{H}}^T)\}$ be the entire \textit{sequence} of robot states and human commands that the robot observed over the course of an interaction \footnote{We emphasize that these interactions can be of any length, and that there is no maximum interaction  length specified to the algorithm.}. As the human and robot repeatedly collaborate and interact, the robot collects a \textit{dataset} of these sequences: $\mathcal{D} = \{\tau^1, \tau^2, \ldots, \tau^{i-1} \}$. Notice that here we distinguish the current interaction $\tau^i$. Because the robot only knows the states and human inputs up to the present time, for the current interaction $\tau^i = \{(s^1, a_{\mathcal{H}}^1), \ldots, (s^{t-1}, a_{\mathcal{H}}^{t-1})\}$.

\p{Robot}  In settings where an assistive robot arm repeatedly interacts with a human the robot has access to four pieces of information. The robot knows its state $s$, the start and end of current interaction $(\tau^i_{start} , \tau^i_{end} \in \tau^i)$, the human's behavior during the current interaction $\tau^i$, and the events of previous interactions $\mathcal{D}$. Given $(s, \tau^i, \mathcal{D})$, the robot needs to decide: (a) what assistance $a_{\mathcal{R}}$ to provide and (b) how to arbitrate control between the human and robot through $\beta$. We emphasize that under this formulation the robot makes no assumptions about either the human's underlying tasks or how to complete them --- instead, the robot must extract this information from previous interactions. In practice, the designer may choose to initialize the robot with some tasks before the assistive arm encounters the current user. Under our formulation this prior information takes the form of offline demonstrations: the designer could provide interactions $\mathcal{D}_{\text{offline}}$, so that $\mathcal{D} \rightarrow \mathcal{D} \cup \mathcal{D}_{\text{offline}}$. Moving forward, the robot must leverage the available data $(s, \tau^i, \mathcal{D})$ to learn to share autonomy with the current human operator.

%% file: method.tex
\section{Learning to Share Autonomy across Repeated Interaction (\textbf{SARI})}\label{sec:method}

\begin{figure*}[t!]
	\begin{center}
		\includegraphics[width=\columnwidth]{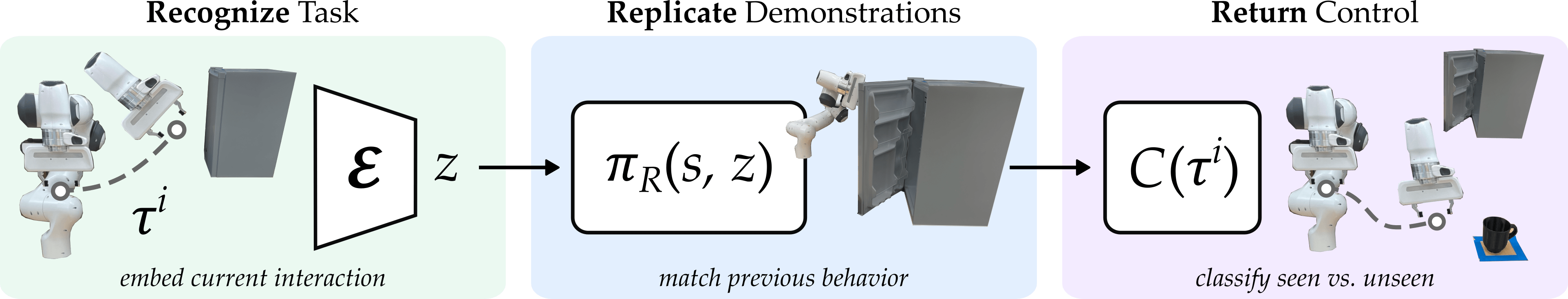}

		\caption{Outline of SARI, our proposed algorithmic framework for learning to share autonomy across repeated interaction. (Left) The robot embeds the human's behavior $\tau^i$ during the current interaction to a distribution over latent tasks $z$. (Middle) The robot then chooses assistive actions $a_{\mathcal{R}}$ conditioned on its state $s$ and latent task $z$. The assistive policy $\pi_{\mathcal{R}}$ is trained to match the user's behavior from previous interactions. (Right) To decide whether or not to trust this assistive action, the robot turns to a discriminator $\mathcal{C}$. The discriminator assesses whether the current interaction $\tau^i$ is similar to any previously seen interaction: if so, the robot increases autonomy. In this example the robot remembers how the human has opened the fridge in the past, and assists for that task. But when the human does something new (reaching for the cup) the robot realizes that it does not know how to help, and arbitrates control back to the human.}
		\label{fig:methods}
	\end{center}
    \vspace{-1em}

\end{figure*}

Our proposed approach is guided by the intuition that --- if the robot recognizes the human's behavior is similar to a previous interaction --- the robot can assist the human by imitating that past interaction. Take our motivating example of opening the fridge door: the next time the human starts guiding the robot towards this door, the robot should infer which task the human is trying to perform and then autonomously open the door just like the human. There are three key challenges to this problem. First, the robot must \textbf{\textit{recognize}} the human's task $z^i$ during the current interaction. Next, the robot should \textbf{\textit{replicate}} any previous interactions that are similar to this task. Finally, the robot must know when it is unsure, and \textbf{\textit{return}} control to the human if the task is new or unexpected. In this section we introduce an algorithm to tackle these three challenges (see \fig{methods}). We refer to our method as \textbf{SARI}: \textbf{S}hared \textbf{A}utonomy across \textbf{R}epeated \textbf{I}nteractions.

\subsection{\emph{Recognize}: Embedding Interactions to a Latent Space} \label{sec:encoder}

Our first step is to extract the user's high-level task $z^i$ from the robot's low-level observations. Recall that the human's behavior during the current interaction (i.e., their commanded actions at each robot state) is captured in $\tau^i$. This behavior is guided by the human's desired task: when the human wants to open the fridge, they provide commands $a_{\mathcal{H}}$ that move the robot towards that door, and when the human wants to pick up a coffee cup, they provide a different set of commands to reach that cup. Accordingly, we leverage $\tau^i$ to recognize the underlying task $z^i$. More formally, we introduce an encoder:
\begin{equation} \label{eq:M1}
    z \sim \mathcal{E}(~\cdot \mid \tau^i)
\end{equation}
This encoder \textit{embeds} the human's behavior into a probability distribution over the latent space $\mathcal{Z} \subseteq \mathbb{R}^d$. We will learn the encoder model from previous human interactions as described in the following subsection (\ref{sec:decoder}).

Our encoder $\mathcal{E}$ is analogous to \textit{goal prediction} from prior works on shared autonomy \cite{dragan2013policy, jain2019probabilistic, javdani2018shared, brooks2019balanced, newman2018harmonic, nikolaidis2017human, aronson2018eye, admoni2016predicting}. In these prior works the robot observes the human's current behavior $\tau^i$, and then applies Bayesian inference to predict the human's goal $z^i$. Our encoder ${\mathcal{E}(z \mid \tau^i)}$ accomplishes the same thing: it outputs a distribution over tasks the human may want to complete. The difference is that --- when using Bayesian inference --- the robot needs to know the set of possible tasks \textit{a priori}. When training the encoder we make no such assumption. Instead, the encoder \textit{learns} a distribution over tasks using only past interactions $\mathcal{D}$. However, one practical concern here is that the robot could convince itself of its own prediction: i.e., because the robot is autonomously moving towards a goal, the robot might think that goal is increasingly likely \cite{javdani2018shared}. We avoid this loop by purposely encoding the sequence $\tau^i$. Since $\tau^i$ only includes the human's action $a_{\mathcal{H}}$ (and not the robot's assistance $a_{\mathcal{R}}$), the robot cannot infer a latent task from its own behavior.

\subsection{\emph{Replicate}: Imitating the Demonstrated Behavior} \label{sec:decoder}

As the human uses their joystick to teleoperate the robot towards the fridge door, we leverage our encoder to recognize the human's task. But what does the robot do once it knows that task? And how do we train the encoder in the first place? We address both issues by introducing a robot policy (i.e., a decoder) that maps our task predictions into assistive robot actions:
\begin{equation} \label{eq:M2}
    a_{\mathcal{R}} = \pi_{\mathcal{R}}(s, z)
\end{equation}
The policy $\pi_{\mathcal{R}}$ determines how the robot assists the human. We want the robot's policy to imitate previous demonstrations, so that if the human's current behavior is similar to another interaction $\tau \in \mathcal{D}$, the robot will generalize the human's actions from that previous interaction.

We accomplish this by training the encoder and policy models using the dataset of previous interactions $\mathcal{D}$. More specifically, we take snippets of the human's behavior during previous interactions, embed those snippets to a task prediction, and then reconstruct the human's demonstrated behavior. For some past interaction $\tau \in \mathcal{D}$, let $\xi = \{(s^1, a_{\mathcal{H}}^1), \ldots (s^{k-1}, a_{\mathcal{H}}^{k-1})\}$ be the human's behavior up to timestep $k$, and let $(s^k, a_{\mathcal{H}}^k)$ be the human's behavior at timestep $k$. We train the encoder and policy to minimize the loss function:
\begin{equation} \label{eq:M3}
    \mathcal{L} = \mathbb{E}_{z \sim \mathcal{E}(\cdot \mid \xi)} ~ \Big\|a_{\mathcal{H}}^k - \pi_{\mathcal{R}}(s^k, z) \Big\|^2
\end{equation}
across the dataset $\mathcal{D}$. In other words, we train the encoder from \eq{M1} and policy from \eq{M2} so that --- given a snippet of the human's past behavior --- we correctly predict the human's next action. \eq{M3} encourages the robot to mimic the human, so that when the robot encounters a familiar task, the arm will take autonomous actions that match the commands which the human previously provided. As a reminder, here the robot is not simply saving and replaying the human's demonstrations: because each interaction is slightly different, the robot is learning a policy model $\pi_{\mathcal{R}}$ to generalize the human's demonstrations to nearby states.

We contrast our policy to \textit{trajectory prediction} \cite{dragan2013policy, jain2019probabilistic} or \textit{task constraints} \cite{bragg2020fake, broad2020data, reddy2018shared, schaff2020residual, losey2022learning} from previous research on shared autonomy. Under these approaches the robot assumes that it knows the correct way to perform each task; e.g., if the human wants to reach for a cup, the robot assumes that it should move in a straight line towards that goal. But we know that tasks are \textit{personalized}, and different users will complete the same task in different ways. Instead of constraining robot assistance to a pre-specified task definition, we therefore learn to imitate how the current user performs each task and train $\pi_\mathcal{R}$ to match the current user's behaviors.

\subsection{\emph{Return}: Knowing What We Do Not Know} \label{sec:classify}

If the human repeats a task that the robot has seen many times before (e.g., opening the fridge), we can rely on our model to assist the human. But what happens if the human tries to perform a new or rarely seen task? Here we \textit{do not trust} the robot's assistive actions since this task is out of the robot's training distribution. In general, deciding where to arbitrate control requires a trade-off: we want the robot to take as many autonomous actions as possible (reducing the human's burden), but we do not want the assistive robot to over-commit to incorrect autonomous actions and prevent the human from doing what they actually intended.

To solve this problem we take inspiration from recent work on interactive and safe imitation learning \cite{zhang2017query,menda2017dropoutdagger,kelly2019hg, hoque2022thriftydagger}. Our objective is to determine when the robot should trust the collective output of Equations~(\ref{eq:M1}) and (\ref{eq:M2}). Intuitively, if the human's behavior $\tau^i$ is unlike any seen behavior $\tau \in \mathcal{D}$ we should return control to the human. We therefore train a discriminator $\mathcal{C}$ that distinguishes \textit{seen} behavior from \textit{unseen} behavior. Unseen behavior is cheap to produce: we can generate this behavior by applying noisy deformations to the observed interactions $\tau \in \mathcal{D}$ \cite{losey2017trajectory}. At run time, our discriminator outputs a scalar confidence over the human's current behavior, which we then utilize to arbitrate control between human and robot:
\begin{equation} \label{eq:M4}
    \beta \propto \mathcal{C}(\tau^i)
\end{equation}
In our experiments we implement this as $\mathcal{C}(\tau^i) = \mathcal{C}(s^t, a^t_{\mathcal{H}})$, where $(s^t, a^t_{\mathcal{H}})$ is the most recent state-action pair from $\tau^i$, and the output of $\mathcal{C}$ is normalized using a softmax function \cite{goodfellow2016} to obtain the arbitration parameter $\beta \in [0, 1]$. Recall that $\beta$ from \eq{P2} blends the robot and human actions $a_{\mathcal{R}}$ and $a_{\mathcal{H}}$. If $\tau^i$ deviates from previously seen input patterns, $\beta \rightarrow 0$, and the robot returns control to the human operator. By contrast, if the discriminator recognizes $\tau^i$ as similar to previous experience, $\beta \rightarrow 1$ and the robot arm partially automates its motion.

\p{Continual Learning} During each interaction the robot applies Equations~(\ref{eq:M1}), (\ref{eq:M2}), and (\ref{eq:M4}) to assist the human. But what about \textit{between} interactions, when the human is not providing any inputs to the robot? Imagine we train our encoder, policy, and discriminator after the human has collaborated with the robot for a few minutes. Over the next hour the human will inevitably perform new tasks. An intelligent assistive robot should also learn these tasks and \textit{continuously adapt} to the human. At the end of interaction $i$, we therefore add $\tau^i$ to dataset $\mathcal{D}$. We then \textit{retrain} SARI between interactions, updating $\mathcal{E}$, $\pi_{\mathcal{R}}$, and $\mathcal{C}$. Intermittent retraining enables the robot to continually learn and refine tasks over long-term interaction.

%% file: theory.tex
\section{Analyzing Stability with SARI} \label{sec:theory}

The SARI algorithm we introduced in Section~\ref{sec:method} learns to recognize tasks, replicate demonstrations, and return control. Here we apply stability theory to this learning approach. Specifically, we explore the performance of SARI when the human attempts to complete a new, previously unseen task. We know that the robot \textit{should} recognize its uncertainty and arbitrate control back to the human. But the robot is also trying to provide assistance and reduce the human's burden --- and if the robot mistakenly thinks it knows the human's intent, our system may override the user and autonomously perform the wrong task. For example, in \fig{methods} this \textit{false positive} causes the robot to open the fridge (a previously seen skill) instead of reaching for the cup (a new and unexpected goal).

Motivated by this failure case, we bound the error between the robot's final state and the human's desired goal. We first start with a single degree-of-freedom system for the sake of clarity, and then extend our analysis to $d$-dimensional robot arms. Overall, we prove that the final state error of a SARI robot --- i.e., the distance between $s$ and the human's goal --- is \textit{uniformly ultimately bounded}. The radius of this bound is a function of the SARI design parameters, the distance between the human's new task and previously seen tasks, and the variance in the human teleoperator's inputs. Throughout this section we conduct experiments with simulated humans and simulated or real robot arms: we find that our theoretical error bounds align with the measured error in these studies. Finally, we extract practical guidelines that designers can leverage to tune the hyperparameters of our SARI algorithm.

\subsection{Single Degree-of-Freedom System}\label{sec:univariate}

To more clearly explain our theoretical analysis we start by considering a $1$-DoF assistive robot. Here the state $s \in \mathbb{R}$, the human command $a_\mathcal{H} \in \mathbb{R}$, and the robot assistance $a_\mathcal{R} \in \mathbb{R}$ are all scalars. The human is teaching this robot to reach for static goals\footnote{Our analysis can also be extended to continuous skills by assuming that the human's goal is the closest waypoint along the skill's trajectory.}. At every previous interaction the human teleoperated the robot towards a \textit{known goal} $g$. Now the human changes their mind and attempts to reach a \textit{new goal} $g^*$. Returning to our motivating example from \fig{methods}, perhaps the human has repeatedly teleoperated their assistive arm to open the fridge, and now at interaction $i$ the user wants to pick up a cup.

\p{Robot} During past interactions the human guided the robot towards $g$. We assume these past human actions were sampled from a Gaussian distribution $a_{\mathcal{H}} \sim \mathcal{N}\big((g-s), \sigma_{\mathcal{D}}^2)\big)$ that nosily moved from $s$ to $g$. Applying SARI, the robot collects these state-action pairs into sequences ${\tau = \{(s^1, a_{\mathcal{H}}^1), \ldots, (s^T, a_{\mathcal{H}}^T)\}}$ and a dataset ${\mathcal{D} = \{\tau^1, \ldots, \tau^{i-1}\}}$. The robot then learns to recognize and replicate the human's behavior by minimizing \eq{M3} across the dataset $\mathcal{D}$. We assume a \textit{best-case} robot that learns to perfectly match the human's past behavior such that the robot's assistive policy is $a_{\mathcal{R}} \sim \mathcal{N}\big((g-s), \sigma_{\mathcal{D}}^2)\big)$. Similarly, this best-case robot learns to return control such that the discriminator --- i.e., $\beta$ in \eq{M4} --- is the robot's learned policy evaluated at the human's current action:
\begin{equation} \label{eq:T1}
    \beta(s, a_{\mathcal{H}})
    = \frac{1}{\sqrt{2\pi\sigma_{\mathcal{D}}^2}}\exp{ \left( \frac{-\big(a_{\mathcal{H}}-(g-s)\big)^2}{2\sigma_{\mathcal{D}}^2}\right)}
\end{equation}
We emphasize that $\sigma_{\mathcal{D}}$ captures the precision and consistency of the human's previous interactions. Here $\sigma_{\mathcal{D}} \rightarrow 0$ indicates that the human directly guided the robot to the known goal $g$, while $\sigma_{\mathcal{D}} \rightarrow \infty$ indicates that the human's past interactions were noisy and imperfect (i.e., the human may have pressed the joystick in the wrong direction or overshot their goal).

\p{Human} During the current interaction the human reaches for a new, unexpected goal $g^*$. As before, we assume the human follows a Gaussian distribution ${a_{\mathcal{H}} \sim \mathcal{N}\big((g^*-s), \sigma_{\mathcal{H}}^2)\big)}$. The standard deviation $\sigma_{\mathcal{H}}$ captures the precision of the human's inputs when reaching for this new goal. We recognize it might be easier (or harder) for the human to teleoperate the robot to the new goal, and thus $\sigma_{\mathcal{H}}$ does not necessarily equal $\sigma_{\mathcal{D}}$.

\p{Lyapunov Stability Analysis} The desired equilibrium of the human-robot system is $s = g^*$, i.e., we want the robot to move to the human's new goal. We propose the Lyapunov candidate function: 
\begin{equation} \label{eq:T2}
    V(t) = \frac{1}{2}e(t)^2, \quad e(t) = g^* - s(t)
\end{equation}
where $e(t) \in \mathbb{R}$ is the error between the robot's state $s$ and the human's goal $g^*$ during the current interaction. Taking the time derivative of \eq{T2}, and substituting in the robot's dynamics from \eq{P1} and \eq{P2}, we obtain:
\begin{equation}\label{eq:T3}
    \dot{V}(t)
    =  - a_{\mathcal{H}}(g^* - s) + \beta a_{\mathcal{H}} (g^* - s) -  \beta a_{\mathcal{R}}(g^*-s)
\end{equation}
Recall that $a_{\mathcal{H}}$, $a_{\mathcal{R}}$, and $\beta$ are all probabilistic quantities. We take the expectation of $\dot{V}$ to reach: 
\begin{equation}\label{eq:T4}
    \mathbb{E}[\dot{V}(t)]
    = -(g^* - s)^2 + \mathbb{E}[\beta a_\mathcal{H}] (g^* - s) - \mathbb{E}[\beta] (g^*-s) (g-s) 
\end{equation}
Intuitively, we want \eq{T4} to be negative so that $V(t)$ \textit{decreases} over time and the human-robot system approaches equilibrium $e(t)=0$ in expectation.

For our next steps it is critical to understand the role of the arbitration factor $\beta$. Recalling that ${a_{\mathcal{H}} \sim \mathcal{N}\big((g^*-s), \sigma_{\mathcal{H}}^2)\big)}$, we take the expectation of \eq{T1} to reach:
\begin{equation}\label{eq:T5}
\mathbb{E}[\beta] =
  \dfrac{1}{\sqrt{2\pi(\sigma_{\mathcal{D}}^2 + \sigma_{\mathcal{H}}^2)}}\exp{ \left( \dfrac{-(g^*-g)^2}{2(\sigma_{\mathcal{D}}^2 + \sigma_{\mathcal{H}}^2)}\right)}
\end{equation}
This function does not always hold. From our original definition in \eq{T2} we remember that $\beta \in [0, 1]$, where $\beta\rightarrow 0$ corresponds to full human control and $\beta\rightarrow 1$ is fully autonomous behavior. In practice, designers may further limit $\beta \leq \beta_{max}$, $\beta_{max} \in (0, 1]$, so that the human maintains a persistent minimal control over the assistive robot \cite{newman2018harmonic, dragan2013policy, selvaggio2021autonomy}. Accordingly, we here reach \textit{two cases} for our stability analysis: (a) when $\mathbb{E}[\beta] \geq \beta_{max}$ and (b) when $\mathbb{E}[\beta] < \beta_{max}$. Below we derive two separate stability results for both of these cases.

\p{Theorem 1}\label{thm:T1}
Consider a $1$-DoF robot using SARI. Given the robot's learned policy is $a_{\mathcal{R}} \sim \mathcal{N}\big((g-s), \sigma_{\mathcal{D}}^2)\big)$, the human's current policy is ${a_{\mathcal{H}} \sim \mathcal{N}\big((g^*-s), \sigma_{\mathcal{H}}^2)\big)}$, and $\mathbb{E}[\beta] \geq \beta_{max}$ in \eq{T5}, the error is \emph{uniformly ultimately bounded}. The ultimate bound is: 
\begin{equation}\label{eq:T6}
    |g^* - s| > \beta_{max} \cdot |g^* - g|
\end{equation}
\smallskip
\noindent \textit{Proof.} Recall that for persistent minimal control we require that $\beta \leq \beta_{max}$. Since we have $\mathbb{E}[\beta] \geq \beta_{max}$ and the system cannot physically exceed $\beta_{max}$,  we set $\beta = \beta_{max}$. Accordingly, we have that $\mathbb{E}[\beta] = \beta_{max}$ and $\mathbb{E}[\beta a_{\mathcal{H}}] = \beta_{max} \cdot \mathbb{E}[a_{\mathcal{H}}]$ in \eq{T4}. Rearranging the updated \eq{T4}, we find that $\mathbb{E}[\dot{V}(t)] < 0$ when \eq{T6} is satisfied. It follows that the human-robot system is uniformly ultimately bounded \cite{spong2006robot, khalil2002nonlinear}. See the Appendix for more details. \qed

\medskip

We can intuitively think of Theorem 1 as a \textit{false-positive} situation: here $\beta = \beta_{max}$ and the robot is fully convinced that the human's current goal is $g$. Fortunately, our SARI algorithm is designed to prevent false-positives by returning control when the robot is faced with new or previously unseen behaviors. This leads to our second setting where $\mathbb{E}[\beta] < \beta_{max}$.

\p{Theorem 2}\label{thm:T2}
Consider a $1$-DoF robot using SARI. Given the same conditions as in Theorem 1, but now $\mathbb{E}[\beta] < \beta_{max}$, the error is \emph{uniformly ultimately bounded}. The ultimate bound is: 
\begin{equation}\label{eq:T7}
    |g^*-s| > \mathbb{E}[\beta] \cdot \frac{\sigma_D^2}{\sigma_{\mathcal{D}}^2+\sigma_{\mathcal{H}}^2} \cdot |g^*-g|
\end{equation}
\smallskip
\noindent \textit{Proof.}
Since $\mathbb{E}[\beta] < \beta_{max}$ we set $\beta = \beta(s, a_{\mathcal{H}})$. We now have that $\beta$ depends upon $a_{\mathcal{H}}$: to compute $\mathbb{E}[\beta a_{\mathcal{H}}]$, we turn to the law of the unconscious statistician (LOTUS) \cite{schervish2014probability}:
\begin{equation}\label{eq:T41}
\mathbb{E}[\beta a_{\mathcal{H}}] = \frac{(g-s)\sigma_{\mathcal{H}}^2 + (g^*-s)\sigma_{\mathcal{D}}^2}{\sqrt{2 \pi}(\sigma_{\mathcal{D}}^2 + \sigma_{\mathcal{H}}^2)^{3/2}} \exp{\left(- \frac{(g^*-g)^2}{2(\sigma_{\mathcal{D}}^2 + \sigma_{\mathcal{H}}^2)}\right)}
\end{equation}
Substituting both \eq{T5} and $\mathbb{E}[\beta a_{\mathcal{H}}]$ back into \eq{T4}, we find that $\mathbb{E}[\dot{V}(t)] < 0$ when \eq{T7} is satisfied. From this it follows that the human-robot system is uniformly ultimately bounded \cite{spong2006robot, khalil2002nonlinear}. \qed

\begin{figure}[t]
	\begin{center}
		\includegraphics[width=1.0\columnwidth]{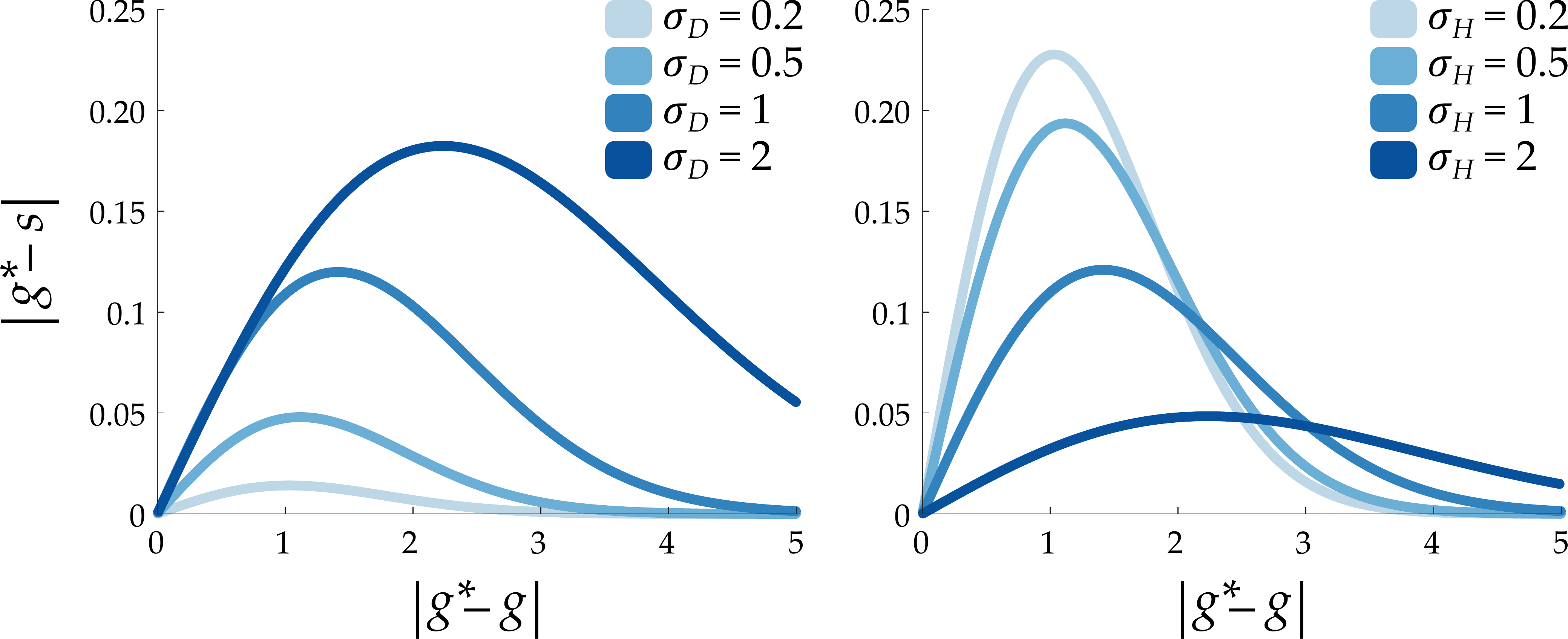}

		\caption{Error bounds for the $1$-DoF system as a function of human noise. All values are in meters. Plots generated using \eq{T6} and \eq{T7} with $\beta_{max} = 1$. (Left) For a fixed $\sigma_{\mathcal{H}}=1$ we increase $\sigma_{\mathcal{D}}$. This captures a human that provided increasingly noisy inputs during past interactions when they were reaching for the known goal $g$. (Right) For a fixed $\sigma_{\mathcal{D}}=1$ we increase $\sigma_{\mathcal{H}}$. This corresponds to a human that provides increasingly noisy inputs during the current interaction while reaching for the new goal $g^*$. We conclude that $\sigma_{\mathcal{D}}$ and $\sigma_{\mathcal{H}}$ have opposite effects on the theoretical error bound.}
		\label{fig:sigma}
	\end{center}

\end{figure}

\p{Implications for SARI} We here highlight \textit{three design guidelines} that emerge from the stability analysis of a $1$-DoF system. First, looking at Theorem 1, we find that lower values of $\beta_{max}$ lead to a decreased error $|g^* - s|$. This aligns with our expectations: when $\beta \rightarrow 0$ the human always retains control and guides the robot without any autonomous intervention. However, smaller values of $\beta_{max}$ also limit the maximum assistance the robot can provide, forcing the human to continually teleoperate the robot arm. Hence, choosing $\beta_{max}$ is a \textit{trade-off} between increased error bounds and increased human effort.

Second, from Theorem 2 the precision of the human's previous interactions $(\sigma_{\mathcal{D}})$ and current interaction $(\sigma_{\mathcal{H}})$ have opposite effects on the error bound (see \fig{sigma}). Humans that accurately moved to goal $g$ will have lower error bounds when reaching for the new goal; i.e., \textit{decreasing} $\sigma_{\mathcal{D}}$ reduces the error $|g^* - s|$. Conversely, after the human starts moving towards the new goal $g^*$, noisy motions are beneficial: \textit{increasing} $\sigma_{\mathcal{H}}$ reduces the error $|g^* - s|$. We tie both of these trends back to the discriminator in \eq{M4}. When the human's inputs are easily distinguished from previous interactions --- i.e., when the human takes actions the robot has not seen before --- SARI returns control and the human can reach their new goal.

Finally, by combining Theorem 1 and Theorem 2 we have that the worst-case error occurs when the human's new goal $g^*$ is \textit{close to} --- but not the exact same as --- the human's previous goal $g$. As $|g^*-g|$ increases $\mathbb{E}[\beta] \rightarrow 0$ and the error bound in \eq{T7} approaches zero. Similarly, when $|g^*-g| \rightarrow 0$ we have that $|g^*-s| \rightarrow 0$ in both \eq{T6} and \eq{T7}. Our results here are consistent with \cite{fontaine2020quality}, where Fontaine \textit{et al.} demonstrate that nearby goals are an adversarial setting for existing shared autonomy algorithms.

\begin{figure}[t!]
	\begin{center}
		\includegraphics[width=1.0\columnwidth]{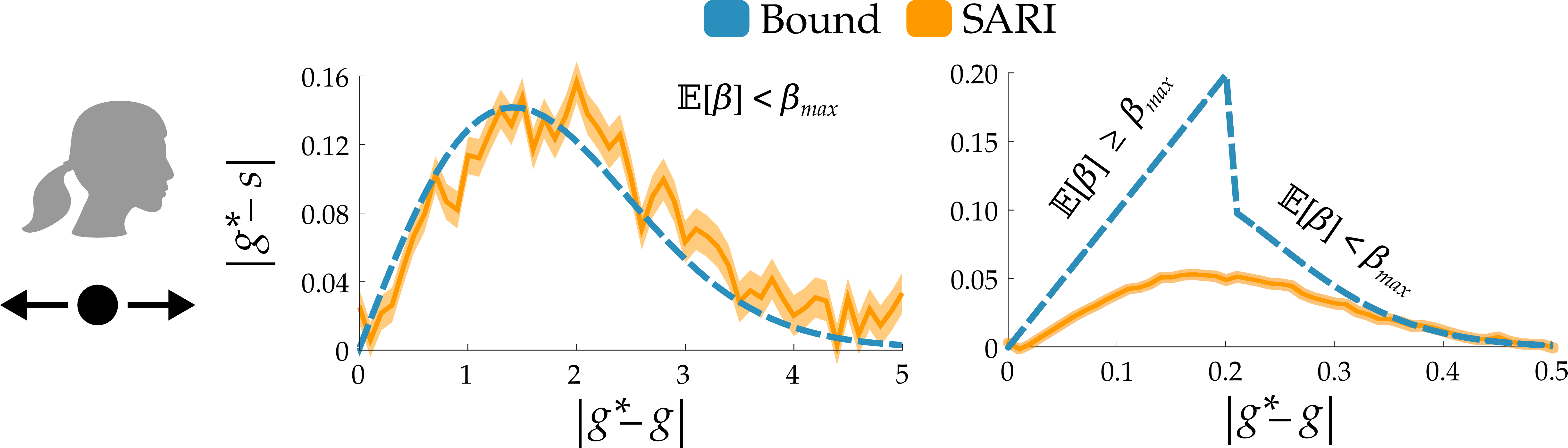}

        \caption{Error bound and experimental results for a $1$-DoF SARI system. All values are in meters. Here a simulated Gaussian human provided $250$ demonstrations reaching for their original goal $g$. These demonstrations were used to train the SARI algorithm; at test time the simulated human reached for a series of new goals $g^*$ with SARI assistance. For each $g^*$ we collected $10,000$ runs --- the shaded region is the standard deviation across these runs. (Center) While reaching for the previous goal $g$ and new goal $g^*$ the human had noise $\sigma_{\mathcal{D}}=\sigma_{\mathcal{H}}=1$. For all choices of $g^*$ we have that $\mathbb{E}[\beta] < \beta_{max}$ in \eq{T5}, and thus the theoretical bound is \eq{T7}. (Right) We choose $\sigma_{\mathcal{D}} = \sigma_{\mathcal{H}}=0.1$ and had two different theoretical error bounds: When $g^*$ is close to $g$ then $\mathbb{E}[\beta] \geq \beta_{max}$ and \eq{T6} applies; but as $g^*$ get farther from $g$ we have that $\mathbb{E}[\beta] < \beta_{max}$, and thus the bound is \eq{T7}. The bound appears tight when $\mathbb{E}[\beta] < \beta_{max}$ and more conservative when $\mathbb{E}[\beta] \geq \beta_{max}$.}
		
		\label{fig:single}
	\end{center}

\end{figure}

\p{Experimental Validation} In our analysis we have made two important assumptions about SARI. First, we assumed that the robot's learned policy $a_{\mathcal{H}} \sim \mathcal{N}\big((g-s), \sigma_{\mathcal{D}}^2)\big)$ exactly replicates the human's previous behavior. Second, we assumed that the robot's discriminator learns \eq{T1}, i.e., the discriminator outputs the likelihood of the human's current action under the robot's learned policy. In \fig{single} we test both of those assumptions by \textit{comparing the theoretical bounds} from \eq{T6} and \eq{T7} to the \textit{experimental behavior} of our SARI algorithm.

To generate these plots we simulated a $1$-DoF point-mass robot and human operator. The simulated human repeatedly guided the robot to a known goal $g$ with actions $a_{\mathcal{H}} \sim \mathcal{N}\big((g-s), \sigma_{\mathcal{D}}^2)\big)$, and we followed the procedure from Section~\ref{sec:method} to train our SARI algorithm on these interactions. The simulated human then takes actions ${a_{\mathcal{H}} \sim \mathcal{N}\big((g^*-s), \sigma_{\mathcal{H}}^2)\big)}$ to reach a new goal $g^*$ while receiving SARI assistance. For values of $g^*$ where Theorem 2 applies (i.e., when $\mathbb{E}[\beta] < \beta_{max}$) we find a close correspondence between \eq{T7} and the robot's measured error $e(t)$. For values of $g^*$ where $\mathbb{E}[\beta] \geq \beta_{max}$ it appears that Theorem 1 becomes overly conservative: the experimental error is consistently lower than \eq{T6}.

\subsection{Multiple Degree-of-Freedom System} \label{sec:multivariate}

So far we have explored the stability and error bounds of a $1$-DoF system. We now extend these results to the general case. Here the state $\boldsymbol{s} \in \mathbb{R}^d$, the human command $\boldsymbol{a_\mathcal{H}} \in \mathbb{R}^d$, and the robot assistance $\boldsymbol{a_\mathcal{R}} \in \mathbb{R}^d$ are all $d$-dimensional vectors. To better distinguish that we are working with vectors we will \textbf{bold} these symbols for this section of the paper. Our problem setup is the same as in Section~\ref{sec:univariate}: during past interactions the human guided the robot towards goal $\boldsymbol{g}$, and we want to evaluate error during the current interaction when the human is teleoperating the robot to a new, previously unseen goal $\boldsymbol{g^*}$. 

\p{Assumptions} As before, we assume that the human's actions during past interactions were sampled from a multivariate Gaussian distribution $\boldsymbol{a_\mathcal{H}} \sim \mathcal{N}\big((\boldsymbol{g} - \boldsymbol{s}), \Sigma_{\mathcal{D}}\big)$. During the current interaction the human noisily moves towards their new goal by following the policy $\boldsymbol{a_\mathcal{H}} \sim \mathcal{N}\big((\boldsymbol{g^*} - \boldsymbol{s}), \Sigma_{\mathcal{H}}\big)$. We make two key assumptions about SARI. (a) Our approach learns to perfectly recognize and replicate the human's past behavior, and provides assistive actions $\boldsymbol{a_\mathcal{R}} \sim \mathcal{N}\big((\boldsymbol{g} - \boldsymbol{s}), \Sigma_{\mathcal{D}}\big)$. (b) Our discriminator learns to output a scalar $\beta$ that matches the robot's policy evaluated at the human's action $\boldsymbol{a_\mathcal{H}}$. These assumptions are the same as in Section~\ref{sec:univariate}. We note that the covariance matrix $\Sigma_{\mathcal{D}}$ captures the human's noise during past interactions, and $\Sigma_{\mathcal{H}}$ is the noise during the current interaction.

\p{Lyapunov Stability Analysis} We want SARI to drive the human-robot system towards the equilibrium $\boldsymbol{s} = \boldsymbol{g^*}$. We accordingly propose the Lyapunov candidate function:
\begin{equation} \label{eq:T8}
    V(t) = \frac{1}{2}\|\boldsymbol{e}(t)\|^2, \quad \boldsymbol{e}(t) = \boldsymbol{g^*} - \boldsymbol{s}(t)
\end{equation}
Taking the time derivative of \eq{T8}, plugging in the robot's dynamics from \eq{P1} and \eq{P2}, and then taking the expectation, we obtain:
\begin{equation}\label{eq:T9}
    \mathbb{E}[\dot{V}(t)] = -\boldsymbol{e}^T  \Big(\boldsymbol{e} - \mathbb{E}[\beta \boldsymbol{a_\mathcal{H}}] + \mathbb{E}[\beta] \boldsymbol(\boldsymbol{g} - \boldsymbol{s})\Big)
\end{equation}
Our goal here is to find a condition that ensures $\mathbb{E}[\dot{V}(t)] < 0$ so that the human-robot system approaches equilibrium $\boldsymbol{e}(t) = 0$. Importantly, the next steps of our analysis depend on the expected value of $\beta$. Given our assumptions about SARI, for a $d$-dimensional system we find that:
\begin{equation}\label{eq:T10}
\mathbb{E}[\beta] =
\frac{1}{\sqrt{(2\pi)^d \det\Sigma}}\exp{\left(-\frac{1}{2}\|\boldsymbol{g^*} - \boldsymbol{g}\|^2_{\Sigma^{-1}}\right)}
\end{equation}
where $\Sigma = \Sigma_{\mathcal{D}} + \Sigma_{\mathcal{H}}$ is the sum of the covariance matrices. Recalling that the arbitration factor $\beta$ must be within $[0, \beta_{max}]$, we again reach two \textit{two cases} for our stability analysis: (a) when $\mathbb{E}[\beta] \geq \beta_{max}$ and (b) when $\mathbb{E}[\beta] < \beta_{max}$. Below we list the general stability results for each case.

\begin{figure}
	\begin{center}
		\includegraphics[width=0.65\columnwidth]{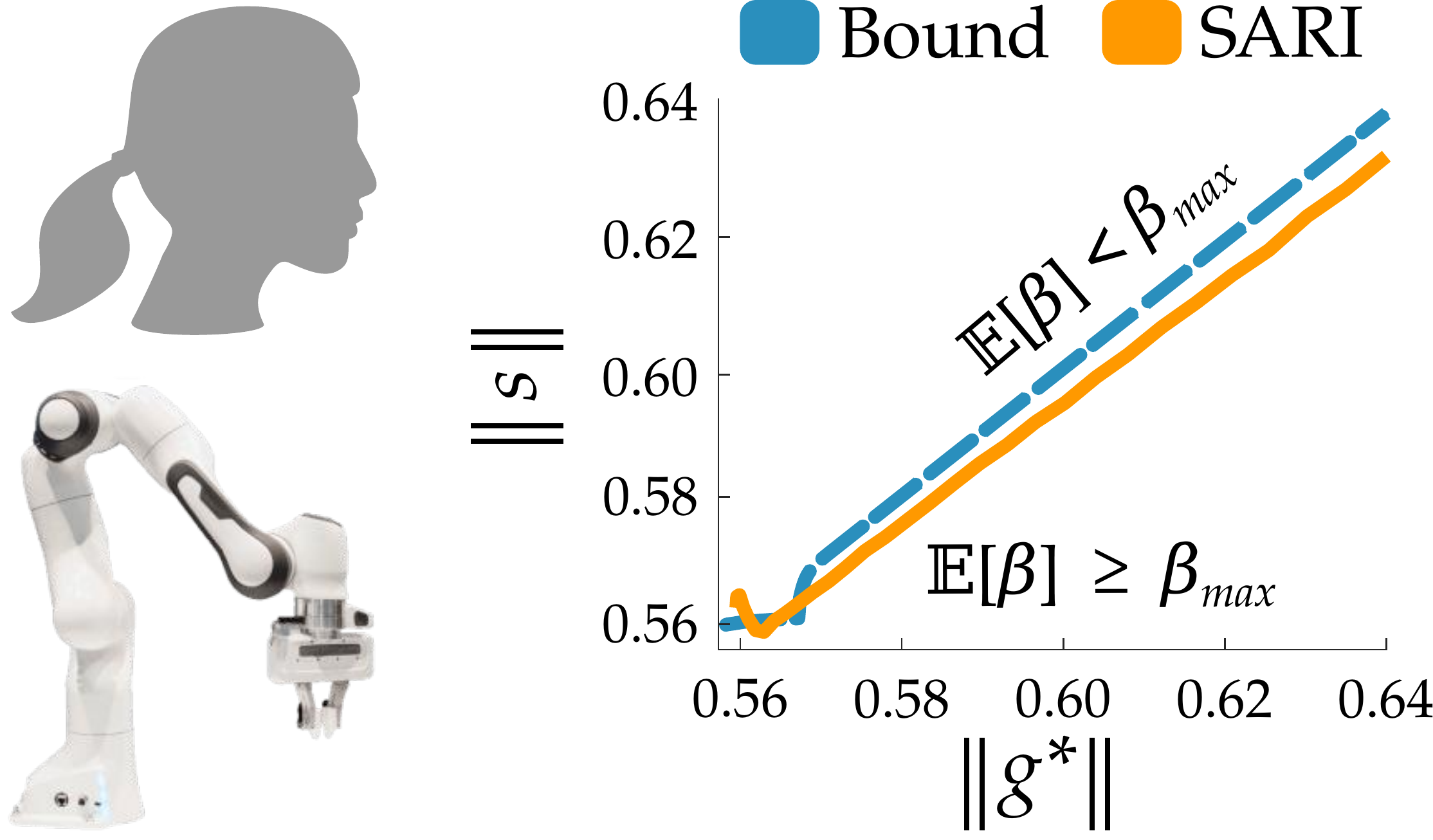}
    
        \caption{Error bound and experimental results for a $3$-DoF SARI system. All values are in meters, and ideally $\|\boldsymbol{s}\| = \|\boldsymbol{g^*}\|$. A simulated Gaussian human provided $25$ demonstrations reaching for their original $(x, y, z)$ goal position on the Franka robot arm. These demonstrations were used to train the SARI algorithm; at test time the simulated human reached a series of new goals $\boldsymbol{g^*}$ with SARI assistance. For each $\boldsymbol{g^*}$ we collected $5$ runs. While reaching for the previous goal $\boldsymbol{g}$ and new goal $\boldsymbol{g^*}$ the human had noise $\Sigma_{\mathcal{D}}=\Sigma_{\mathcal{H}}=1e^{-4} \cdot I$, where $I$ is the identity matrix. For choices of $\|\boldsymbol{g^*}\|$ close to $0.56$, we have that $\mathbb{E}[\beta] \geq \beta_{max}$ and the bound is given by \eq{T11}. As $\|\boldsymbol{g^*}\|$ increases beyond $0.57$, we find $\mathbb{E}[\beta] < \beta_{max}$ and the bound is \eq{T12}.}

		\label{fig:theo2}
	\end{center}

\end{figure}

\smallskip

\p{Theorem 3}
Consider a $d$-DoF robot using SARI. Given the robot's learned policy is $\boldsymbol{a_\mathcal{H}} \sim \mathcal{N}\big((\boldsymbol{g} - \boldsymbol{s}), \Sigma_{\mathcal{D}}\big)$, the human's current policy is $\boldsymbol{a_\mathcal{H}} \sim \mathcal{N}\big((\boldsymbol{g^*} - \boldsymbol{s}), \Sigma_{\mathcal{H}}\big)$, and $\mathbb{E}[\beta] \geq \beta_{max}$ in \eq{T10}, the error is \emph{uniformly ultimately bounded}. The ultimate bound is: 
\begin{equation}\label{eq:T11}
      \|\boldsymbol{g^*} - \boldsymbol{s}\| > \beta_{max} \cdot \|\boldsymbol{g}^* - \boldsymbol{g}\|
\end{equation}
\smallskip
\noindent \textit{Proof.}
Since $\mathbb{E}[\beta] \geq \beta_{max}$ we set $\beta = \beta_{max}$. Hence $\mathbb{E}[\beta] = \beta_{max}$ and $\mathbb{E}[\beta \boldsymbol{a_\mathcal{H}}] = \beta_{max} (\boldsymbol{g}^* - \boldsymbol{s})$. Substituting this into \eq{T9} and applying the Cauchy–Schwarz inequality, $\mathbb{E}[\dot{V}(t)] < 0$ when \eq{T11} is satisfied. It follows that the human-robot system is uniformly ultimately bounded \cite{spong2006robot, khalil2002nonlinear}. See our Appendix for more details. \qed

\p{Theorem 4}
Given a $d$-DoF SARI robot under the same conditions as in Theorem 3, but now with $\mathbb{E}[\beta] < \beta_{max}$, the error is \emph{uniformly ultimately bounded}. The ultimate bound is:
\begin{equation}\label{eq:T12}
    \|\boldsymbol{g^*}-\boldsymbol{s}\| > \lambda \mathbb{E}[\beta] \cdot \|\boldsymbol{g^*} - \boldsymbol{g}\|
\end{equation}
where $\lambda$ is the maximum eigenvalue of $(\Sigma_{\mathcal{D}} + \Sigma_{\mathcal{H}})^{-1}\Sigma_D$.

\medskip

The full proof for Theorem 4 can be found in the Appendix. We highlight that if $d=1$ and we have a single DoF robot, then \eq{T11} and \eq{T12} are equivalent to our univariate results from \eq{T6} and \eq{T7}. Overall, the SARI error bounds are a function of the designer's choice of $\beta_{max}$, the amount of noise in the operator's joystick inputs, and the distance between the previous and new goals.

\p{Experimental Validation} To support our stability analysis we compared the theoretical error bounds from \eq{T11} and \eq{T12} to the actual behavior of our SARI algorithm. We conducted this study on a Franka Emika Robot arm with a simulated human teleoperator.

The results are shown in \fig{theo2}. The simulated human used a Gaussian policy when reaching for $\boldsymbol{g}$, and we trained SARI using the state-action pairs collected from these interactions. SARI then assisted the simulated human as they reached towards a previously unseen goal $\boldsymbol{g^*}$. We observe a close correspondence between \eq{T12} and the measured error when $\mathbb{E}[\beta] < \beta_{max}$. For goal positions where $\mathbb{E}[\beta] \geq \beta_{max}$ we find that \eq{T11} is conservative, and the actual error is less than our theoretical bound. Viewed together, our results from Sections~\ref{sec:univariate} and \ref{sec:multivariate} support our stability analysis, and suggest that SARI correctly returns control when the human reaches for new and unexpected goals.

%% file: simulations.tex
\section{Simulations} \label{sec:sims}

We have introduced an algorithm that learns to assist users over repeated interaction. Our algorithm (SARI) breaks down into three parts: recognizing the task, replicating prior demonstrations, and returning control when uncertain. In this section we perform simulations to determine how each component of SARI contributes to its overall performance. We recognize that --- in practice --- human operators will use SARI to assist for multiple everyday tasks. Accordingly, we also test the capacity of our approach, and evaluate how SARI's performance changes as it encounters an increasing number of goals and skills. Throughout this section we compare our approach to state-of-the-art imitation learning baselines that also learn from repeated human-robot interaction. We conduct these experiments on both simulated and real robot arms with simulated human operators.

\p{Experimental Setup} For different simulations we implement SARI on either a $7$-DoF Franka Emika robot or a $6$-DoF Universal Robots UR10 robot. We test with two different arms to show that our approach is not hardware specific. A simulated user controls the robot to reach \textit{discrete goals} (e.g., grasping a can) and perform \textit{continuous skills} (e.g., opening a drawer). This simulated user is not perfect: the user selects commanded actions $a_{\mathcal{H}}$ with varying levels of Gaussian white noise, similar to the noisy human models from Section~\ref{sec:theory}. Please see the Appendix for additional details on our implementation.

\subsection{Do We Need Recognition?} \label{sec:sim1}

\begin{figure}[t!]
	\begin{center}
		\includegraphics[width=0.75\columnwidth]{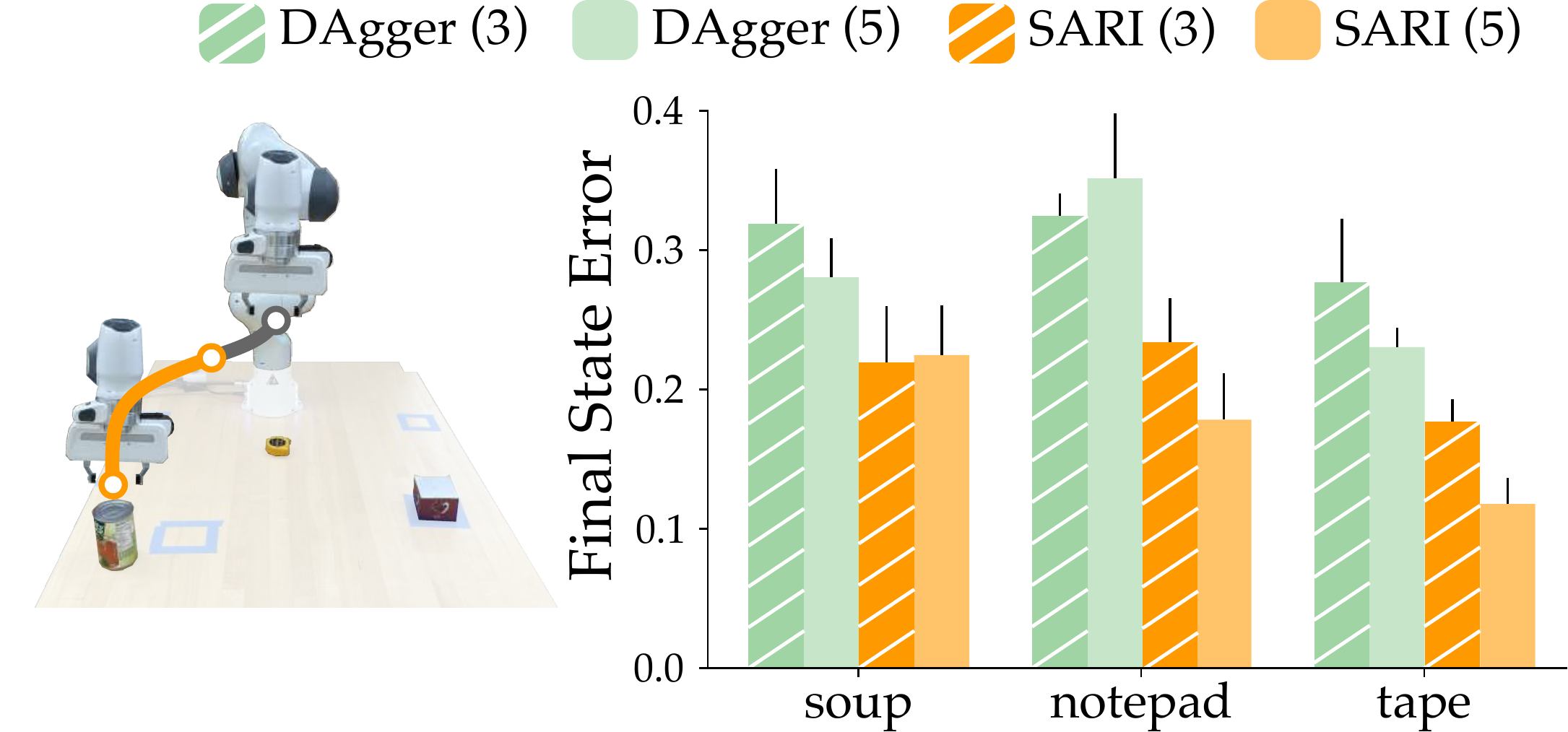}

		\caption{Comparison to \textbf{DAgger} \cite{ross2011reduction}, an imitation learning baseline that does not use latent embeddings. A simulated user controls the robot for the first $0.5$ seconds of the interaction: the robot must recognize the human's task and complete the rest of the reaching motion autonomously. We measure the final state error for each goal after training with $3$ or $5$ repeated interactions. Comparing all \textbf{DAgger} runs to all \textbf{SARI} runs, we find that the final state error is lower with \textbf{SARI}: $t(29)=3.215$, $p<0.05$.}
		
		\label{fig:sim1}
	\end{center}
	\vspace{-1em}

\end{figure}

In our first experiment we explore whether we need two separate modules for task recognition and replication. Recall that in Section~\ref{sec:encoder} we introduced an encoder which embeds the current interaction $\tau^i$ into a latent task prediction $z \in \mathcal{Z}$. Within Section~\ref{sec:decoder} we then mapped $z$ to an assistive robot actions using $\pi_{\mathcal{R}}(s, z)$. Here we test whether we need this encoder in the first place: in other words, can we obtain similar performance \textit{without} embedding to latent space $\mathcal{Z}$? We consider an imitation learning baseline that \textit{directly} maps the current interaction $\tau^i$ to robot actions $a_{\mathcal{R}}$ using a learned policy $\pi_{\mathcal{R}}(s, \tau^i)$. Specifically,
we compare \textbf{SARI} against \textbf{DAgger} \cite{ross2011reduction}.

This experiment was performed on the Franka Emika robot arm with a simulated human (see \fig{sim1}). The environment consisted of three potential goals: a can of soup, a notepad, or a tape measure. The human first teleoperated the robot along $3$ or $5$ demonstrations to reach each goal. We trained \textbf{SARI} and \textbf{DAgger} from these repeated interactions such that both approaches had access to the same training data. At test time, the human guided the robot for the first $0.5$~s of the task: based on this input, the robot had to (a) recognize which task the human was trying to perform and (b) automate the rest of the reaching motion. We plot the resulting error between the human's goal and the robot's final state in \fig{sim1}. Comparing the results when trained with $3$ or $5$ previous interactions, we find that the robot is better able to provide assistance after additional interactions. Regardless of whether \textbf{DAgger} had $3$ or $5$ demonstrations, however, \textbf{SARI} more accurately reached the human's goal given the same simulated human operator. We verify the significance of these results using a paired
t-test ($t(29) = 3.215, p<0.05$) over $5$ separate trials with \textbf{DAgger} and \textbf{SARI}. These results suggest that incorporating a separate encoder for task recognition \textit{improves} the robot's assistance.

\begin{figure}[t!]
	\begin{center}
		\includegraphics[width=0.75\columnwidth]{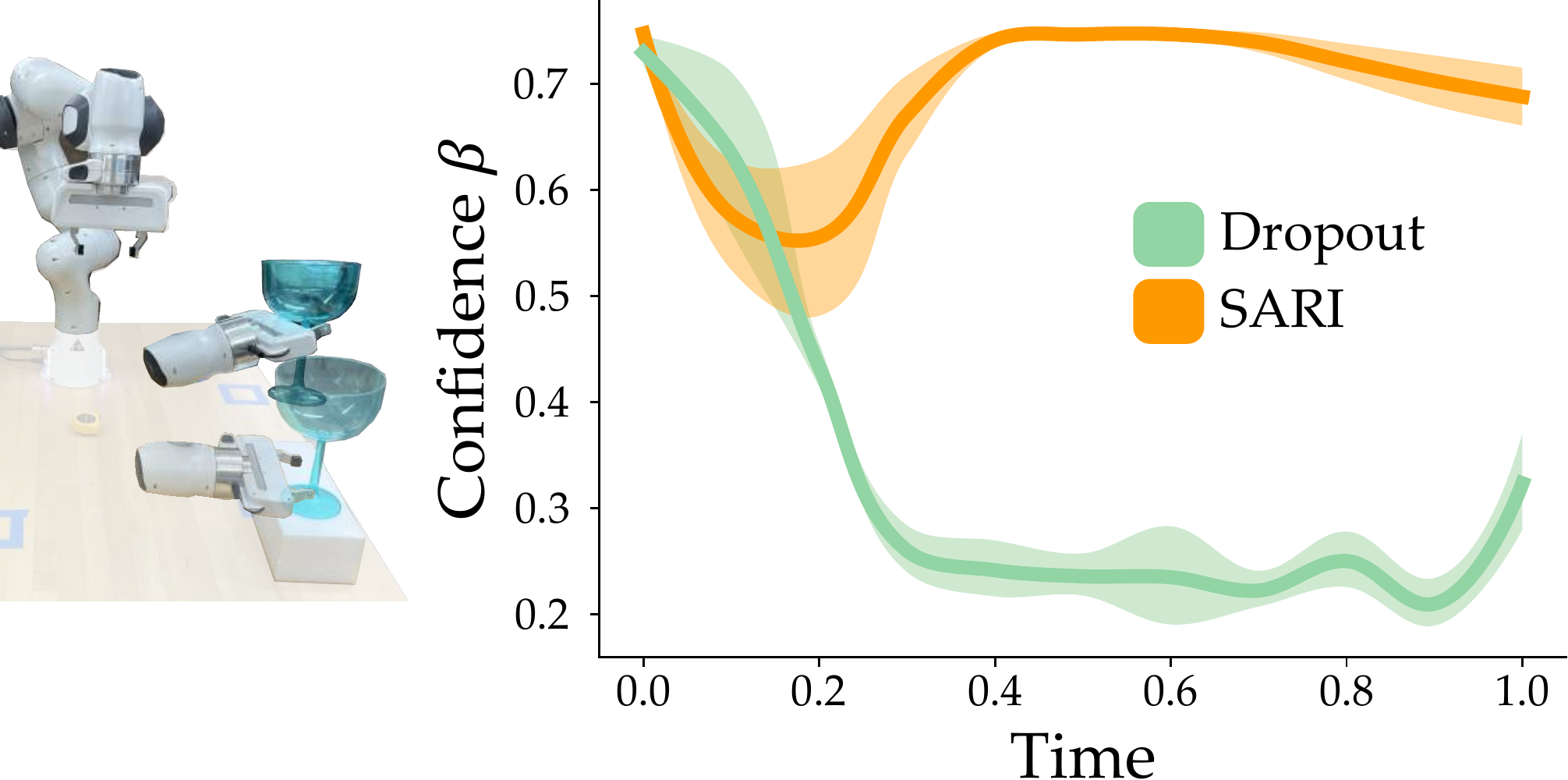}

		\caption{Comparison to \textbf{Dropout}DAgger \cite{menda2017dropoutdagger}, a safe imitation learning baseline where the robot's learned policy $\pi_{\mathcal{R}}$ evaluates its own confidence. Simulated users attempt to lift a glass. Although the robot has seen this continuous skill $5$ times before, with \textbf{Dropout} the robot is overly sensitive to minor deviations from previous interactions and rarely provides assistance.}
		
		\label{fig:sim2}
	\end{center}
	\vspace{-1em}

\end{figure}

\subsection{Do We Need Help Returning Control?}\label{sec:sim_return}

In our second experiment we explore the opposite end of our pipeline: determining when the robot should provide assistance. The stability analysis from Section~\ref{sec:theory} indicates that SARI will return control to the human when the operator is attempting to perform a new task. However, it is equally important for the robot to \textit{retain control} (and provide assistance) when it encounters a \textit{known} task. Here we test if the robot will correctly recognize a previously seen skill. Recall that SARI decides whether or not to provide assistance based on the output of the discriminator from Section~\ref{sec:classify}: this discriminator detects if the state-action pairs in $\tau^i$ are similar to previous interactions. Instead of training a separate discriminator, one alternative is to rely on the confidence of our learned policy \textit{itself}. Here we turn to prior work on safe imitation learning where the robot samples its learned policy multiple times at the current state, and assesses the similarity of the resulting actions $a_{\mathcal{R}}$. If all of these actions are almost identical, the robot is \textit{confident} it knows what to do; conversely, if the model outputs have high variance, the robot is \textit{unsure}. We therefore compare \textbf{SARI} to DropoutDAgger \cite{menda2017dropoutdagger} (\textbf{Dropout}). 

This experiment was performed on a Franka Emika robot arm with a simulated human teleoperator (see \fig{sim2}). The simulated human and real robot attempted to complete a continuous manipulation task where the robot must reach and lift a glass. During test time, the human and robot shared control throughout the entire interaction using \eq{P2}. The robot had seen the human perform this task in five past interactions, and so it should have been confident when providing assistance. To ensure proper comparison, we train \textbf{SARI} and \textbf{Dropout} on $5$ separate trials. We visualize the robot's mean actual confidence $\beta$ over these trails in \fig{sim2}. Interestingly, we find that \textbf{Dropout} is overly sensitive to minor deviations from previous interactions, and incorrectly returns control to the human even when the robot can still provide useful assistance. \textbf{SARI} remains confident throughout this known task, suggesting that our separate discriminator better arbitrates control than the learned policy itself. This is important in practice: human operators will never perform the same task in the exact same way, and thus the robot must be able to remain confident on known tasks despite some operator variability.

\subsection{What if the Operator is Increasingly Noisy?}

So far we have focused on the robot's perspective, and have tested each component of our SARI approach. For our third experiment we instead focus on the \textit{human}, and explore how the behavior of the human operator affects SARI. We consider simulated humans with different levels of noise as they attempt to complete previously seen or new tasks. For previously seen tasks, we want to make sure that the robot continues to provide assistance even as the human becomes an increasingly noisy and imperfect operator. For new tasks, we recognize that the robot should not \textit{resist} the human or \textit{force} them along a previously seen trajectory. This problem becomes particularly challenging when the human is noisy, since the robot must determine whether the imperfect human is trying to repeat the known task or complete a new task.

\begin{figure}[t!]
	\begin{center}
		\includegraphics[width=1.0\columnwidth]{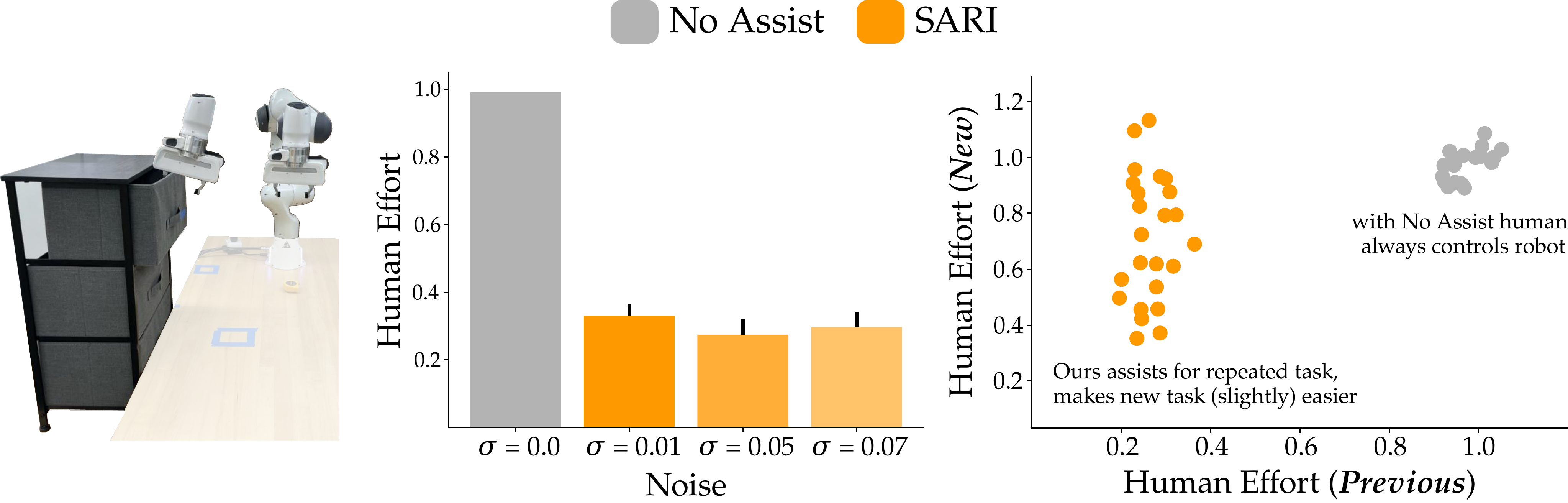}

		\caption{Simulated humans with increasingly noisy behavior. (Center) The human is always attempting to open the drawer (a previous seen skill). We find that \textbf{SARI} correctly recognizes and assists for this task despite noisy and imperfect human teleoperation inputs: $a_{\mathcal{H}} \sim \mathcal{N}\big(a_{\mathcal{H}}^*, \text{diag}(\sigma^2, \ldots, \sigma^2)\big)$. (Right) Simulated users alternate between a previously seen task (opening the drawer) and a new task (reaching a cup). With \textbf{No Assist} both new and previous tasks take about the same amount of Human Effort. \textbf{SARI} learns to partially automate the previously seen task without resisting humans when they try to complete the new task. }
		
		\label{fig:sim31}
	\end{center}

\end{figure}

This experiment was performed on a Franka Emika robot arm with a simulated human teleoperator (see \fig{sim31}). The human provided inputs $a_{\mathcal{H}}^*$ to optimally complete the task, and we then injected Gaussian white noise with covariance matrix $\Sigma_{\mathcal{H}} = \text{diag}(\sigma^2, \ldots, \sigma^2)$. The environment contained a previously seen skill (opening a drawer) and a new goal (reaching a cup). \textbf{SARI} was trained with $5$ repeated interactions of the drawer skill; however, \textbf{SARI} had no prior experience with the cup goal. We compared our approach to a \textbf{No Assist} baseline where the human directly teleoperated the robot's end-effector without any shared autonomy. Consistent with prior experiments, we repeat this experiment $5$ times and measure the mean \textit{Human Effort} across these trials. We define Human Effort as the amount of time the human teleoperates the robot (i.e., the total time the human is providing joystick inputs) normalized by the average time required to complete the task. Lower values of Human Effort signify that the robot correctly automated the motion, while higher values mean that the human had to teleoperate the robot throughout the task.

We first tested the previously seen drawer skill with increasing levels of human noise $\sigma$ (see \fig{sim31}, left). Interestingly, we found that \textbf{SARI} consistently reduced Human Effort while remaining robust to this range of $\sigma$. We then had the simulated human alternate between the new and previous tasks while varying the amount of Gaussian white noise (see \fig{sim31}, right). As expected, \textbf{SARI} made it easier for the human to repeatedly open the drawer --- but on the new task, \textbf{SARI} also correctly returned control back to the human. Performing the new task took no more effort than the \textbf{No Assist} baseline; indeed, it often required \textit{less} human effort. To explain this result, we note that the start of the cup task was similar to the start of the drawer skill, and thus \textbf{SARI} could automate the beginning of this motion (resulting in less Human Effort).

\subsection{How Many Tasks Can We Learn?} \label{sec:sim_capacity}

For our final experiment with simulated humans we explore \textbf{SARI}'s capacity to learn goals and skills. Remember that our motivating application is an assistive robot arm for everyday use: over long-term interaction, this robot will encounter many repeated tasks for which it should provide assistance. More formally, the robot observes an increasing number of interactions $\tau$ and aggregates a growing dataset $\mathcal{D}=\{\tau^1, \ldots, \tau^i\}$. Here we test the performance of SARI as it is trained on this iteratively increasing dataset. We separate the experiment into two parts: in the first environment the robot encounters an increasing number of \textit{goals}, and in the second setting the robot must learn to assist for an increasing number of \textit{skills}. Ideally the SARI robot will have the capacity to learn assistance for \textit{all} of these new tasks, without forgetting or failing to assist for previously seen tasks.

\begin{figure*}[t]
	\begin{center}
		\includegraphics[width=0.75\columnwidth]{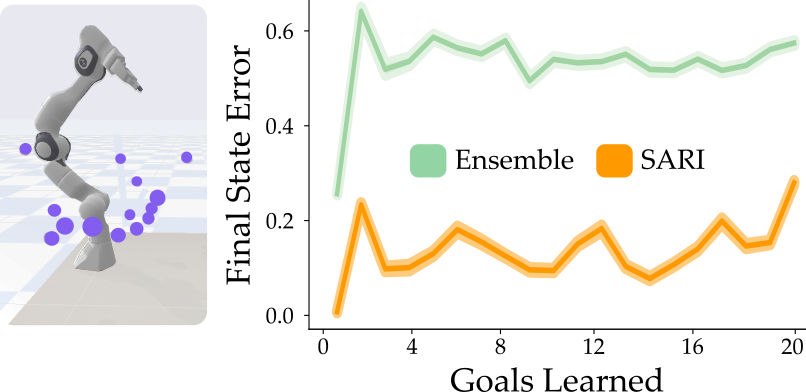}

		\caption{Capacity to assist for an increasing number of goals. A simulated user repeatedly reached for up to $20$ randomly generated goals. The simulated human then teleoperates the robot to reach for all of the goals it has seen so far using two different methods: \textbf{Ensemble} \cite{menda2019ensembledagger} and \textbf{SARI}. \textbf{SARI} has separate modules to recognize, replicate, and return, while \textbf{Ensemble} trains multiple replicate modules and returns control when these policies disagree. Final state error is the difference between the robot's final state and the human's intended goal; the shaded region is the standard error about the mean. \textbf{SARI} maintains roughly constant performance as the number of goals increases, and consistently has lower error than \textbf{Ensemble}.}
		
		\label{fig:sim41}
	\end{center}

\end{figure*}

In both parts of this experiment we compare our proposed approach (\textbf{SARI}) to \textbf{Ensemble}DAgger \cite{menda2019ensembledagger}. \textbf{Ensemble} is an interactive imitation learning approach that trains multiple policies on the human's dataset. Unlike SARI --- which has separate models to recognize, replicate, and return --- under \textbf{Ensemble} the robot only learns policies $\pi_{\mathcal{R}}(s, \tau^i)$ that map the human's behavior directly to robot assistance. The robot trains an ensemble of these policies and compares their outputs: when the actions $a_{\mathcal{R}}$ of each policy agree the robot is confident (i.e., higher $\beta$), and when the actions have high variance the robot is uncertain (i.e., lower $\beta$).

\p{Goals} To explore our method's capacity to learn goals we simulated a Franka Emika robot arm and human operator in PyBullet \cite{coumans2021} (see \fig{sim41}). At the start of the experiment the simulated human repeatedly reached for a single goal, and the robot learned to assist for that goal. Next, the human repeatedly reached for two goals (with new goal positions that were randomly generated), and we tested the robot's ability to assist for \textit{both} goals. Following this pattern, the human iteratively reached for up to $20$ goals; during each iteration we tested the robot's ability to assist for \textit{all the goals} it had observed. This procedure ensures that we are capturing the robot's performance on previously seen goals and the new goal after training. To standardize our results, we trained $20$ separate \textbf{SARI} and \textbf{Ensemble} models at every iteration. Each individual \textbf{SARI} model assisted the simulated human for a single goal $5$ times. For \textbf{Ensemble}, we had the ensemble of $20$ models assist the human $100$ times; put another way, both methods reached for a given goal $100$ total times.

To understand how accurately the human-robot system reached goals, we measured the \textit{Final State Error} between the human's actual goal and the robot's final state. Our results are shown in \fig{sim41}. Overall, we observe that both \textbf{SARI} and \textbf{Ensemble} are \textit{constant} as the number of goals increases: e.g., the error after learning $10$ goals is similar to the error after learning all $20$ goals. But while both approaches have the capacity to learn multiple goals, we find that \textbf{SARI} results in lower error across the board. Our results here are consistent with Section~\ref{sec:sim1}, and suggest that the recognize and return modules in \textbf{SARI} lead to improved performance.

\begin{figure}[t]
	\begin{center}
		\includegraphics[width=0.75\columnwidth]{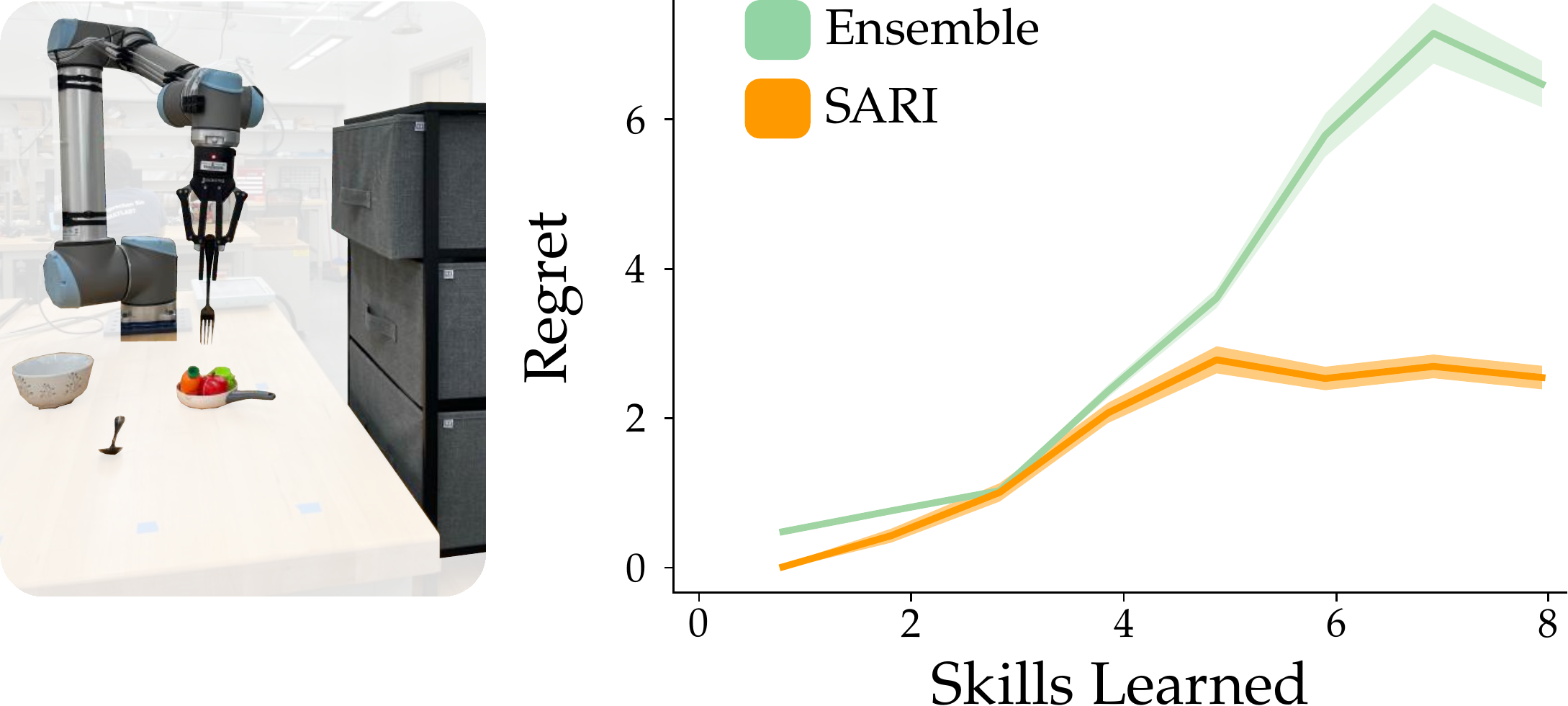}

		\caption{Capacity to assist for an increasing number of skills. A simulated Gaussian human teleoperated the UR10 robot to perform up to $8$ different kitchen skills (e.g., opening a drawer, stabbing a fruit, pushing a bowl). The simulated human then completed these skills when assisted by either \textbf{Ensemble} \cite{menda2019ensembledagger} or \textbf{SARI}. Regret is the difference between the maximum possible reward and the robot's actual reward: lower values of regret indicate the human-robot system completed the skills correctly. Shaded region indicates standard error about the mean. Unlike \fig{sim41}, we observe that performance decreases as the number of skills increases.}
		
		\label{fig:sim42}
	\end{center}

\end{figure}

\p{Skills} To explore our method's capacity to learn continuous skills we paired a simulated Gaussian human with a real $6$-DoF UR10 arm. Here the simulated human attempted to perform kitchen tasks such as opening a drawer, stabbing a piece of fruit, or pushing a bowl (see \fig{sim42}). Similar to \textbf{Goals}, we followed an iterative process: first the human and robot repeatedly performed one skill, then two skills, and so on. At each iteration we evaluated the robot's ability to assist for \textit{all the skills} it had seen so far. We had a total of $8$ skills, and to remove any ordering bias we repeated the experiment twice: once while observing skills $1 \rightarrow 8$, and once while observing skills $8 \rightarrow 1$. To standardize our results, we trained \textbf{SARI} and \textbf{Ensemble} $20$ separate times at each iteration, and evaluated each model's performance $5$ times per skill. 

We measured \textit{Regret} to understand if human-robot system performed each skill correctly. Let $R^*(\xi)$ be the maximum reward that the system can achieve on skill $\xi$; we define \textit{Regret} as $R^*(\xi) - R_{actual}(\xi)$, i.e., the difference between the best-case reward and the robot's actual reward. Our results are plotted in \fig{sim42}. Unlike \textbf{Goals}, we find that the system's performance decreases as the number of skills increases. There are two reasons for this: (a) skills are more complicated than goals, and require more assistance than just a straight point-to-point motion and (b) the robot encounters similar states when performing different skills. This could lead to confusion: if the robot observes state $s$ when pushing the bowl and stabbing the fruit, it is unclear which task the human is currently attempting to perform (and what assistance the robot should provide). Despite these challenges, \textbf{SARI} maintains consistently lower regret when compared to the \textbf{Ensemble} baseline. Overall, our results from both \textbf{Goals} and \textbf{Skills} suggest that \textbf{SARI} has the capacity to learn assistance for multiple tasks. We recognize that this assistance may degrade as the robot continues to aggregate new demonstrations, particularly for skills.

%% file: userstudy.tex
\section{User Studies} \label{sec:user}

In Sections~\ref{sec:theory} and \ref{sec:sims} we studied the theoretical and practical performance of SARI with \textit{simulated} human users. In this section we now turn to user studies with \textit{actual} participants. Recall that our target application is assistive robot arms: we want to enable these arms to share autonomy during everyday tasks. Motivated by this application, we conducted two in-person user studies with \textit{non-disabled} participants and one pilot study with a \textit{disabled} adult who regularly operates assistive robot arms. To show that our method is not specific to particular hardware, we include two different robot arms in our experiments. In the first study participants teleoperated a $7$-DoF Franka Emika robot arm, and in the second and third studies they teleoperated a $6$-DoF Universal Robots UR10 robot arm. The non-disabled participants used a handheld joystick interface for teleoperation (see Figures~\ref{fig:front} and \ref{fig:user1}), while the disabled participant used a web-based interface (see \fig{user3_1}). Additional implementation details for our user studies can be found in the appendix. Video of our user studies is available here: \url{https://youtu.be/3vE4omSvLvc}.

\subsection{Learning Discrete Goals and Continuous Skills} \label{sec:user1}

\begin{figure}[t]
	\begin{center}
		\includegraphics[width=0.75\columnwidth]{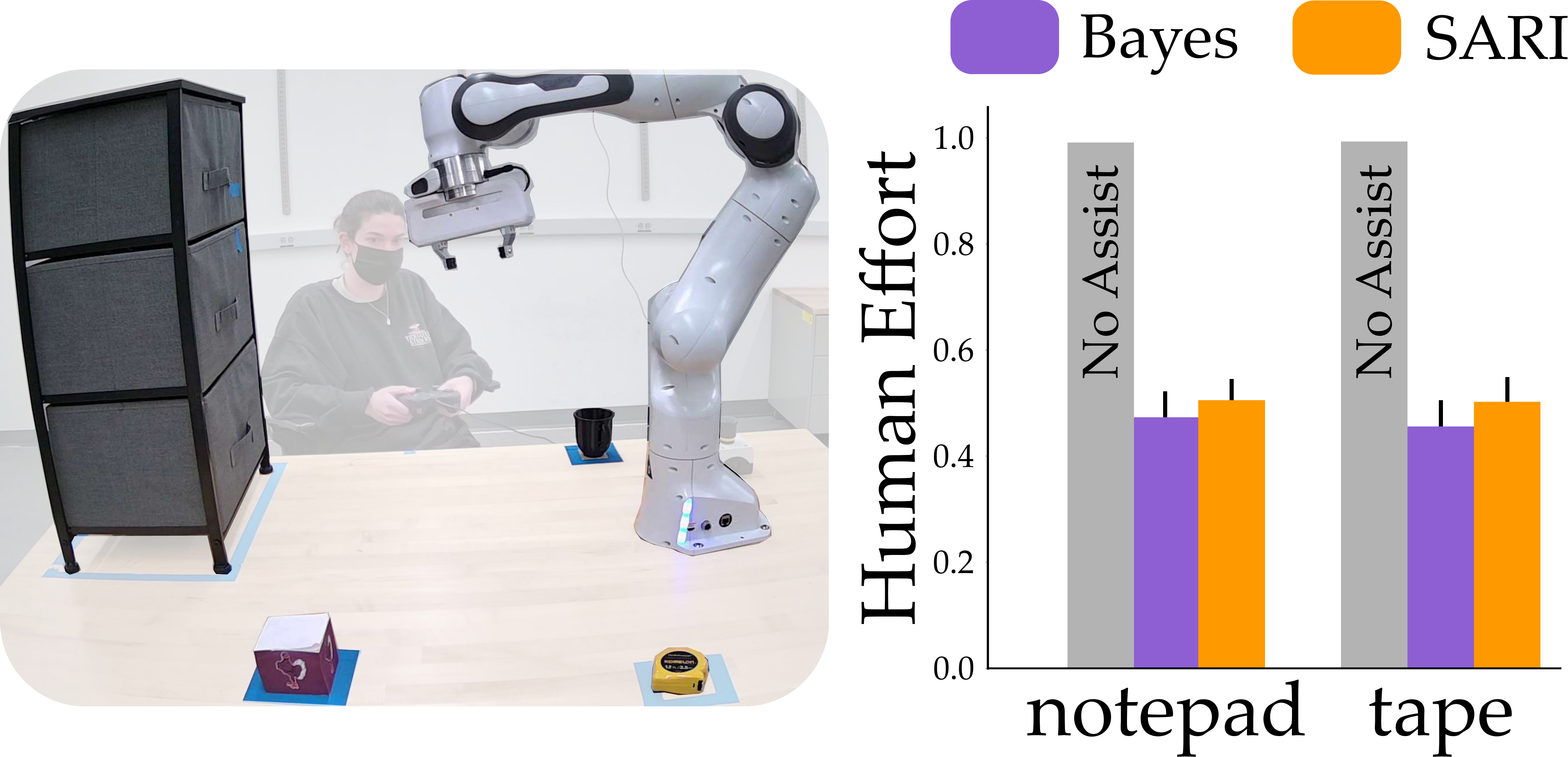}

		\caption{Experimental setup and objective results from the first part of our first user study. Here participants teleoperated the Franka Emika robot arm to reach for two goals that were known \textit{a priori}. We compared our approach (\textbf{SARI}) to a shared autonomy baseline (\textbf{Bayes} \cite{dragan2013policy}) and an industry standard mapping for assistive arm teleoperation (\textbf{No Assist} \cite{kinova}). When the robot has prior knowledge about the human's potential tasks, we find that \textbf{SARI} learns to offer assistance on par with \textbf{Bayes}, and both methods reduce the human's effort when compared to \textbf{No Assist}. Note that \textbf{Bayes} fails to provide helpful assistance when the human wants to perform new, unexpected tasks (e.g., reaching the cup or opening the drawer), as shown in \fig{appendix}.} 
		\label{fig:user1}
	\end{center}

\end{figure}

\begin{figure*}[t]
	\begin{center}
		\includegraphics[width=\columnwidth]{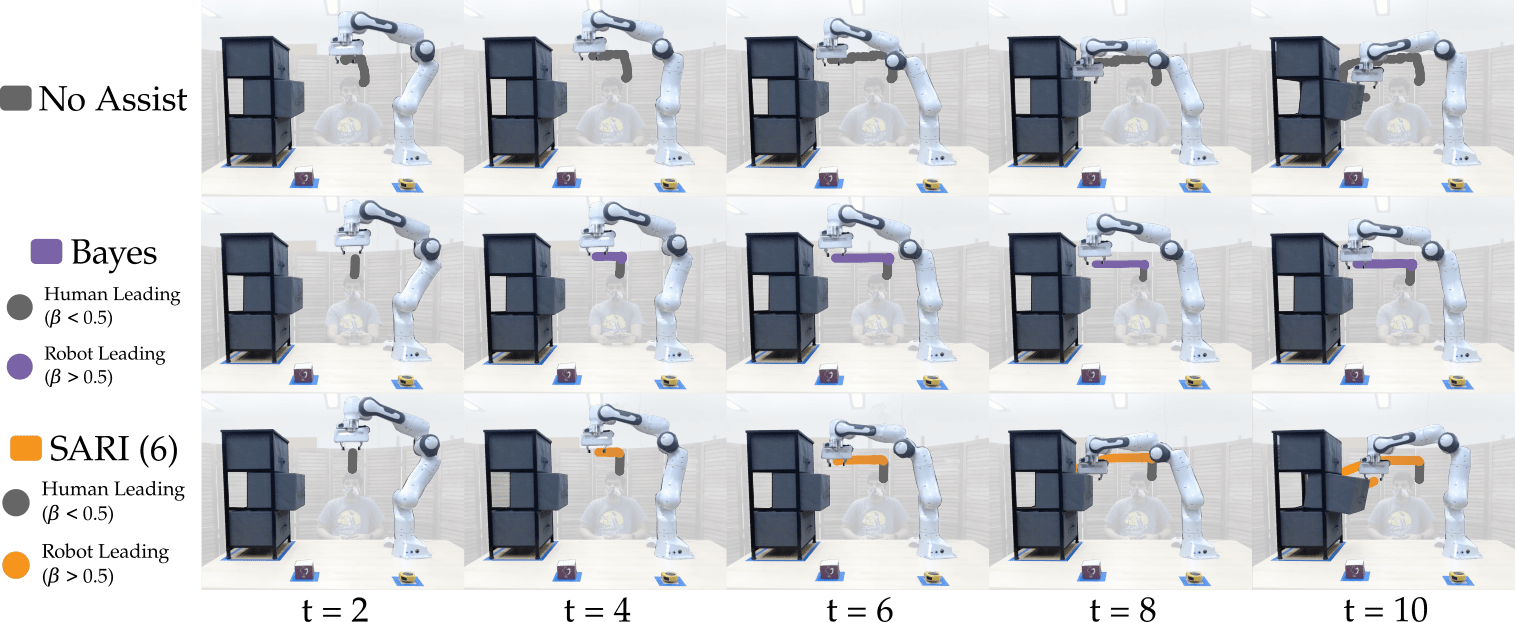}

		\caption{Representative failure case for an existing shared autonomy approach that relies on prior knowledge (\textbf{Bayes} \cite{dragan2013policy}). \textcolor{gray}{\textbf{Gray}} circles indicate the human leading the robot, while \textcolor{violet}{\textbf{purple}} and \textcolor{orange}{\textbf{orange}} indicate the the robot is providing assistance. The user attempts to complete the drawer task with end-effector control (\textbf{No Assist}), with \textbf{Bayes}, and with \textbf{SARI}. None of the methods have prior knowledge about the drawer task; \textbf{Bayes} only knows about the notepad and tape goals, and \textbf{SARI} has observed six repeated interactions for the drawer task. The user is able to successfully open the drawer by themselves (top) and with our method (bottom). With \textbf{SARI} we see that the user is initially leading the robot towards the drawer, but once the robot recognizes this task, it takes charge and offers appropriate assistance (\textcolor{orange}{\textbf{orange}} circles). By contrast, \textbf{Bayes} (middle) mistakes the initial trajectory as towards the notepad, and continually tries to guide the robot to this known goal. Since both the drawer and the notepad are in front of the robot, the robot is initially able to move in the correct direction. However, after the user's inputs diverge from the notepad and go towards the drawer, the robot gets stuck due to conflicting commands.}
		\label{fig:appendix}
	\end{center}
	
\end{figure*}

\begin{figure*}[t!]
	\begin{center}
		\includegraphics[width=\columnwidth]{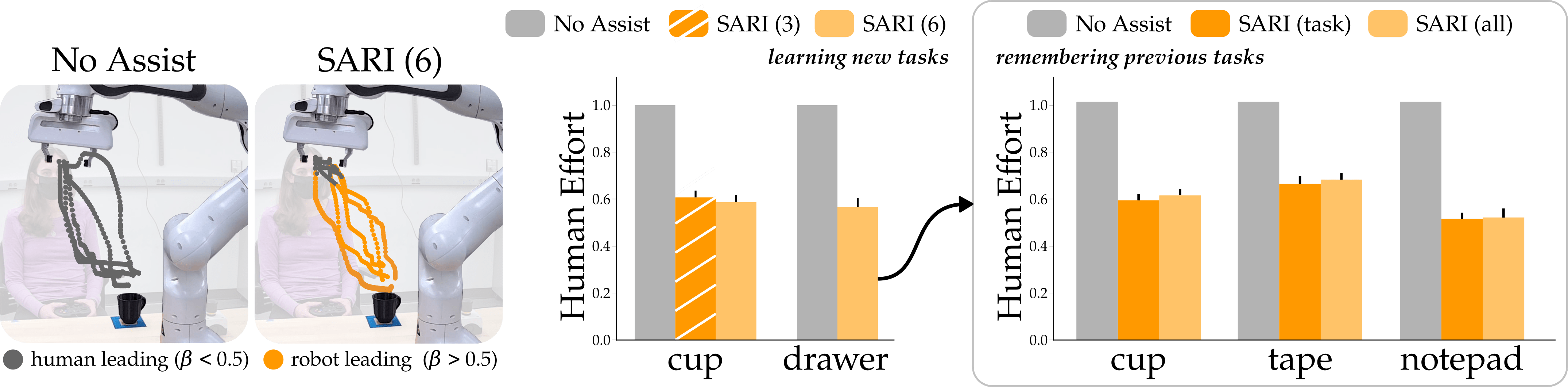}

		\caption{Objective results from the second and third parts of our user study. (Left) The human teleoperates the robot to reach for a goal it did not know about beforehand. The first few times they interact, the user must lead the robot throughout the entire task. After training \textbf{SARI} on six repeated interactions, the robot recognizes the human's intent and automates the rest of the motion; by contrast, with \textbf{No Assist} the human always has to teleoperate the robot. (Center) Across $3-6$ repeated interactions the robot learns to provide assistance for a new goal (reaching a cup) and skill (opening a drawer). This assistance reduces the human's effort as compared to completing the task alone. (Right) We take our resulting model trained on all user demonstrations and revisit the original tasks. \textbf{SARI (all)} offers similar assistance to \textbf{SARI (task)}, a version of our approach trained only with the user's task-specific data. These results suggest our \textbf{SARI} robot has the capacity learn assistance for new tasks without forgetting older ones.}
		\label{fig:user2}
	\end{center}

\end{figure*}

\begin{figure}[t]
	\begin{center}
		\includegraphics[width=0.75\columnwidth]{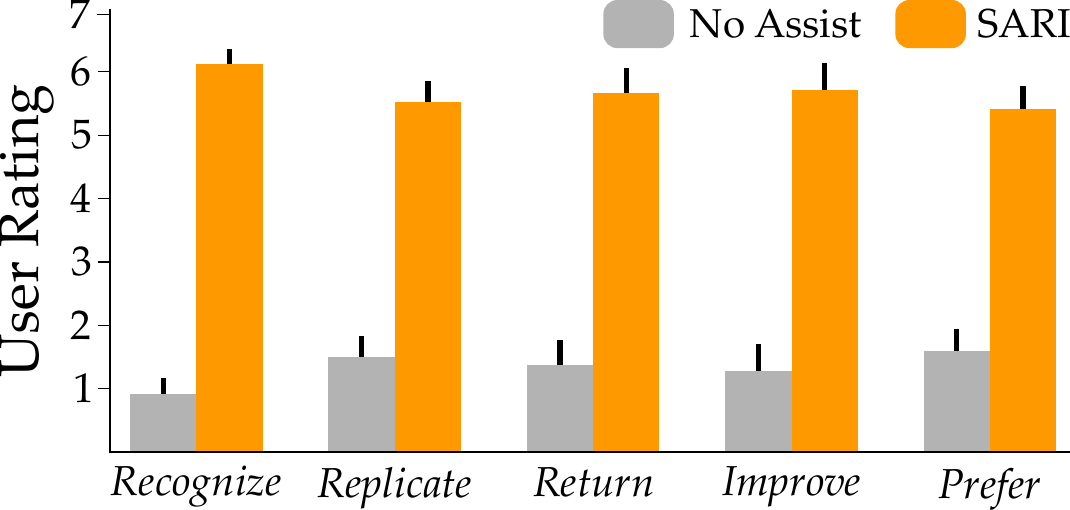}

		\caption{Subjective results from our in-person user study. Higher ratings indicate user agreement. Overall, participants thought \textbf{SARI} provided useful assistance, and they preferred this assistance to trying to complete the tasks with direct end-effector teleoperation (\textbf{No Assist}). These scores were provided after participants had completed the entire experiment, including: completing known tasks, assisting for new tasks, and remembering old tasks.}
		\label{fig:likert}
	\end{center}

\end{figure}

We start with a \textit{three part} user study that explores known and new tasks as well as discrete goals and continuous skills. Non-disabled participants started by reaching for known goals, then taught the robot new skills, and finally returned to the original tasks\footnote{For video of the first user study, also see: \url{https://youtu.be/Plh4t3wQeIA}}. Figures~\ref{fig:user1}, \ref{fig:appendix}, \ref{fig:user2}, and \ref{fig:likert} correspond to this study. We compared our approach (SARI) to two different state-of-the-art baselines: (a) direct end-effector teleoperation that is used on commercial assistive robot arms \cite{jaco, kinova} and (b) an existing shared autonomy algorithm that infers the human's goal from a discrete set of options \cite{dragan2013policy}.

\p{Independent Variables and Experimental Setup} Our first user study was divided into the three sections described below. Each participant completed every section.

In the first part of the user study participants teleoperated the robot to reach for two discrete goals placed on the table. These potential goals were known \textit{a priori}, and the robot had prior experience reaching for them. Here we compare our proposed approach (\textbf{SARI}) to an existing shared autonomy baseline (\textbf{Bayes}) \cite{dragan2013policy}. For \textbf{Bayes} we gave the robot prior information about the location of each goal; during interaction the robot inferred which goal the human wanted and provided assistance towards that goal. For \textbf{SARI} we repeatedly teleoperated the robot to both goals during previous, offline interactions (collecting dataset $\mathcal{D}_{\text{offline}}$). We then trained \textbf{SARI} on this dataset; during interaction the robot used our approach to recognize the user's goal and assist the human for that task.

The shared autonomy baseline is the gold standard when the human wants to complete a task the robot already knows --- but what happens during new tasks? In the second part of our user study participants iteratively performed two new tasks a total of $9$ times each. One task was a discrete goal (reaching a cup), while the other was a continuous skill (opening a drawer). Here we compare \textbf{SARI} to a \textbf{No Assist} baseline. \textbf{No Assist} is direct end-effector teleoperation, and is an industry standard approach for assistive robot arms (e.g., pressing right on the joystick causes the robot to move right) \cite{kinova}. The \textbf{No Assist} baseline never learns from interactions; but for \textbf{SARI} we retrained our approach every three trials during both tasks. We expect that \textbf{SARI} should increasingly assist the user as it gets more familiar with these new tasks.

One concern with our approach is that --- as the robot continues to encounter new tasks --- it will specialize in just one or two recent tasks without remembering how to share autonomy for older tasks. Accordingly, in the last part of the user study participants take the final learned model from both new tasks and use it to revisit the original reaching tasks. Here we compare three conditions: \textbf{No Assist}, where the human acts alone, \textbf{SARI (task)}, the robot's learned assistance with just the user's data from that specific task, and \textbf{SARI (all)}, our approach trained on the user's full dataset of all interactions.

\p{Dependent Measures -- Objective} Across all three parts of the user study we measured \textit{Human Effort}. Human effort is the total time the human teleoperated the robot during the task divided by the average time taken to complete the task. Higher values of human effort indicate that the human had to guide the robot throughout its entire motion, and lower values indicate that the robot partially automated the task.

\p{Dependent Measures -- Subjective} We administered a 7-point Likert scale survey after users completed the study (see Figure \ref{fig:likert}). Questions were organized along five scales: how confident users were that the robot \textit{Recognized} their objective, how helpful the robot's behavior was (\textit{Replicate}), how trustworthy users thought the robot was (\textit{Return}), whether the robot improved after successive demonstrations (\textit{Improve}), and if they would collaborate with the robot again (\textit{Prefer}).

\p{Participants and Procedure}
A total of $10$ members of the Virginia Tech community participated in our study ($3$ female, $1$ non-binary, average age $22 \pm 7$ years). All participants provided informed written consent prior to the experiment under Virginia Tech IRB $\#20$-$755$. 

\p{Hypotheses}
We tested three main hypotheses:
\begin{itemize}
    \item[] \textbf{H1.} \textit{In cases where the robot has prior knowledge about the human's potential goals, SARI will perform similarly to a shared autonomy baseline}
    \item[] \textbf{H2.} \textit{In cases where the human repeatedly performs new and previously unseen tasks, SARI will learn to provide meaningful assistance from scratch}
    \item[] \textbf{H3.} \textit{SARI remembers how to assist users on previously seen tasks even after learning new ones}
\end{itemize}

\p{Results} The results from each part of our user study are visualized in Figures~\ref{fig:user1}, \ref{fig:appendix}, \ref{fig:user2}, and \ref{fig:likert}. In the rest of this subsection we summarize our main findings.

In the first part of the user study participants completed a reaching task with \textbf{Bayes} (a shared autonomy baseline) and \textbf{SARI} (our proposed approach). Here both methods had prior information about the potential goals: for \textbf{Bayes} the robot was given both goal positions, and for \textbf{SARI} we recorded offline interactions reaching for each goal. During the user study the robot had to recognize which goal the human was reaching for (i.e., either the notepad or tape) and then assist the user while reaching for that target. Our results are shown in \fig{user1}. To analyze these results we first performed a repeated measures ANOVA, and found that the robot's algorithm had a significant effect on human effort (Notepad: $F(2, 58)=106$, $p<.001$; Tape: $F(2, 58)=36.9$, $p<.001$). Post hoc comparisons revealed that both \textbf{Bayes} and \textbf{SARI} led to less human effort than \textbf{No Assist}, but the differences between \textbf{Bayes} and \textbf{SARI} were not statistically significant (Notepad: $p=.370$; Tape: $p=.203$). These results suggest that users could reach for known, discrete goals just as easily with \textbf{SARI} as they could with the shared autonomy baseline. 

So \textbf{SARI} is on par with \textbf{Bayes} when the human wants to perform a \textit{known} task --- what happens when the human wants to complete a \textit{new}, unexpected task? To highlight one shortcoming of state-of-the-art shared autonomy approaches and explain why \textbf{Bayes} is not a baseline in the second and third parts of our user study, we illustrate a new task in \fig{appendix}. Here the user attempted to open the drawer, but the robot only had prior knowledge about the notepad and the tape. Recall that under \textbf{Bayes} the robot infers which discrete goal the human is trying to reach and then assists towards that goal \cite{dragan2013policy, jain2019probabilistic, javdani2018shared}. But in this scenario the robot \textit{does not know beforehand} that the human may want to open the drawer. As a result, \textbf{Bayes} misinterpreted the user's inputs and gradually became convinced that the human's target was actually the notepad \textit{next} to the drawer. This interaction ended in a deadlock: the human constantly teleoperated the robot towards the drawer, while the robot continually resisted and refused to return control. Note that the trajectories for \textbf{SARI} and \textbf{No Assist} are similar --- the main difference is that in \textbf{SARI} the robot takes the lead and automates the continuous skill. Moving forward we will focus on new tasks, and will compare \textbf{SARI} to industry standard teleoperation mappings (\textbf{No Assist}).

In the second part of our user study participants repeatedly teleoperated the robot to perform new tasks. This includes the drawer skill in \fig{appendix} and the cup goal in \fig{user2}. During the first few interactions \textbf{SARI} returned control and the user guided the robot throughout the entire task. But after training \textbf{SARI} on $3$ and $6$ repeated interactions, the robot was able to recognize and partially automate these new tasks. One user commented that ``\textit{by the end I didn't provide any assistance and the robot continued to move in the correct direction}.'' We emphasize that --- throughout our entire user study --- the robot was \textit{never told} what task the participant wanted to do. Instead, the robot had to recognize the participant's current task based on that user's joystick inputs. Our results from \fig{user2} suggest that \textbf{SARI} got better at providing assistance for new tasks over repeated interactions. For example, in the drawer skill the human's effort was significantly less with \textbf{SARI} after $6$ repeated interactions ($t(29)=10.5$, $p<.001$).

In the final step of the user study we tested the capacity of our approach. We compared \textbf{SARI} trained on all previous interactions to \textbf{SARI} trained only on interactions for the given task (see \fig{user2}). Intuitively, we expected that the more specialized \textbf{SARI (task)} would provide the best possible performance: this method has only seen data for the current task and therefore cannot misinterpret the human's inputs. Our results suggest that \textbf{SARI} can maintain this performance even when trained with multiple tasks. For three separate goals tasks (cup, tape, and notepad) we first conducted repeated measures ANOVAS, and found that the robot's algorithm had a significant effect on human effort (Cup: $F(2, 38)=51.9$, $p<.001$; Tape: $F(2, 38)=47.7$, $p<.001$; Notepad: $F(2, 38)=74.1$, $p<.001$). Although our approach consistently outperformed \textbf{No Assist}, differences between \textbf{SARI (task)} and \textbf{SARI (all)} were not statistically different (Cup: $p=.416$; Tape: $p=.876$; Notepad: $p=.792$). We therefore conclude that --- similar to our simulations in \fig{sim41} --- \textbf{SARI} has the capacity to learn assistance for new goals without forgetting how to share autonomy on previously seen tasks.

Taken together, these results support \textbf{H1}, \textbf{H2}, and \textbf{H3}. Our approach leveraged repeated interactions to learn to share autonomy across new and old tasks that included discrete goals and continuous skills. Participants generally perceived the robot's assistance as helpful. Looking at the subjective results from \fig{likert}, users thought the robot correctly recognized their intent, made the task easier to complete, and got better at providing assistance over the course of the study.

\subsection{Offering Meaningful Assistance and Returning Control} \label{sec:user2}

\begin{figure*}[t]
	\begin{center}
		\includegraphics[width=0.9\columnwidth]{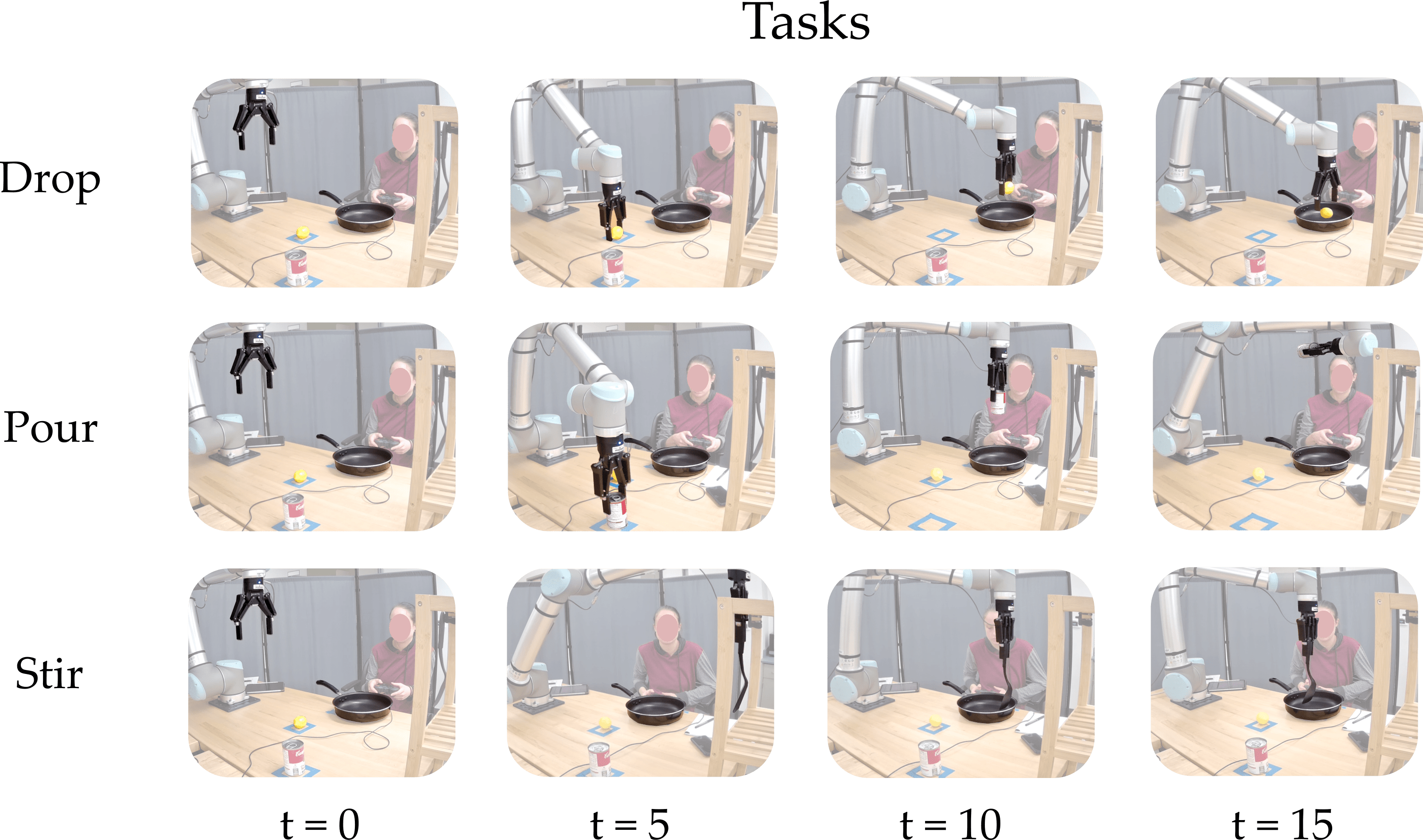}

		\caption{Three tasks used in the second user study. (Top) During \textbf{Drop} users picked up a lemon from the table, moved it to the center of the pan, and then dropped it into the pan. (Middle) During \textbf{Pour} users picked up a can of soup, moved it to the center of the pan, and then poured the soup in the pan. (Bottom) During \textbf{Stir} users fetched a spatula that was placed on a hook, moved towards the top of the pan, and then stirred the pan in a circular motion. All three tasks are continuous skills and were not broken into smaller subtasks. Users completed each task with all three methods (\textbf{No Assist}, \textbf{CASA}, and \textbf{SARI}).}
		\label{fig:user2_1}
	\end{center}

\end{figure*}

\begin{figure*}[t]
	\begin{center}
		\includegraphics[width=0.9\columnwidth]{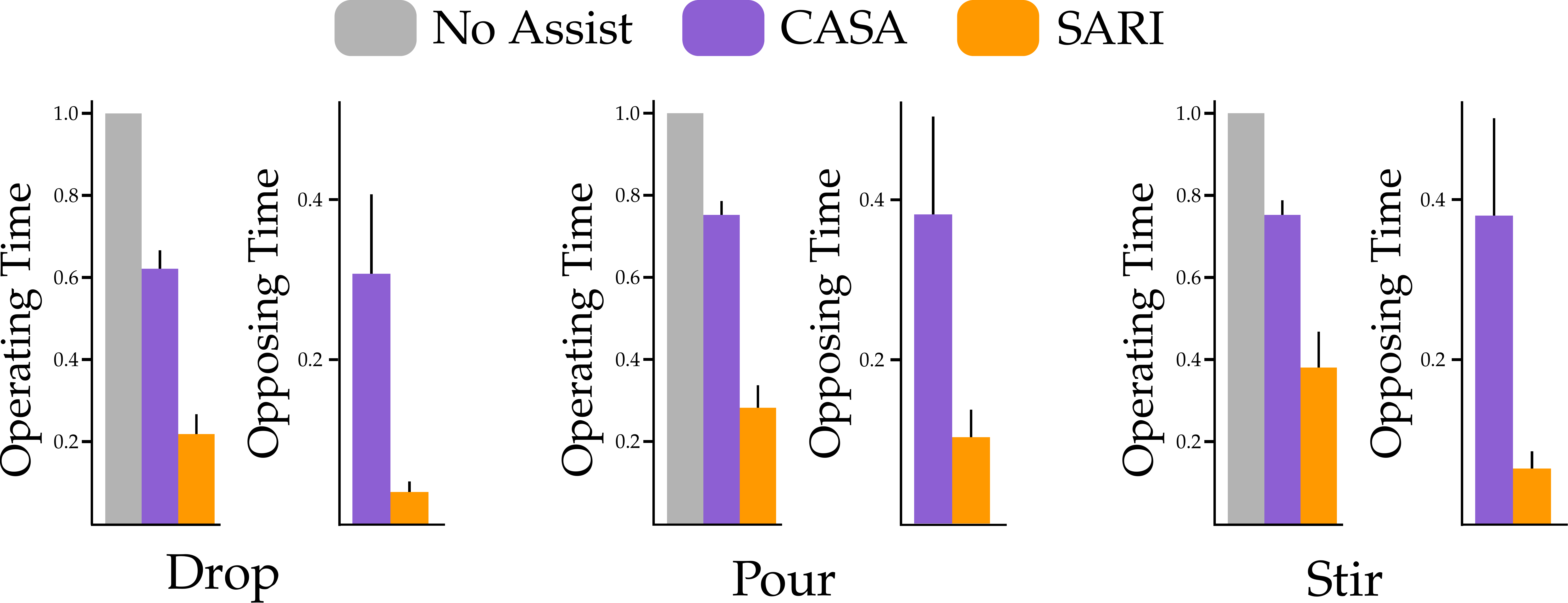}

		\caption{Objective results for the \textbf{known} tasks in our second user study. \textit{Operating Time} is the fraction of the total time where the human provides input to the robot, and \textit{Opposing Time} is the fraction of total time where the robot's actions are antagonistic to the human's input. Note that \textbf{No Assist} has no \textit{Opposing Time} since the in this method the human is always in control of the robot. We find that \textbf{SARI} offers the most meaningful assistance for all three known tasks by reducing both the \textit{Operating Time} and \textit{Opposing Time} for the human. We performed a one way ANOVA analysis and found that the differences between \textbf{CASA} and \textbf{SARI} for both \textit{Operating Time} and \textit{Opposing Time} are significant (Drop: $F(2,87) = 368.6, p<.001$; Pour: $F(2,87) = 347.4, p<.001$; Stir: $F(2,87) = 349.6, p<.001$).}
		\label{fig:user2_2}
	\end{center}

\end{figure*}

\begin{figure*}[t]
	\begin{center}
		\includegraphics[width=0.75\columnwidth]{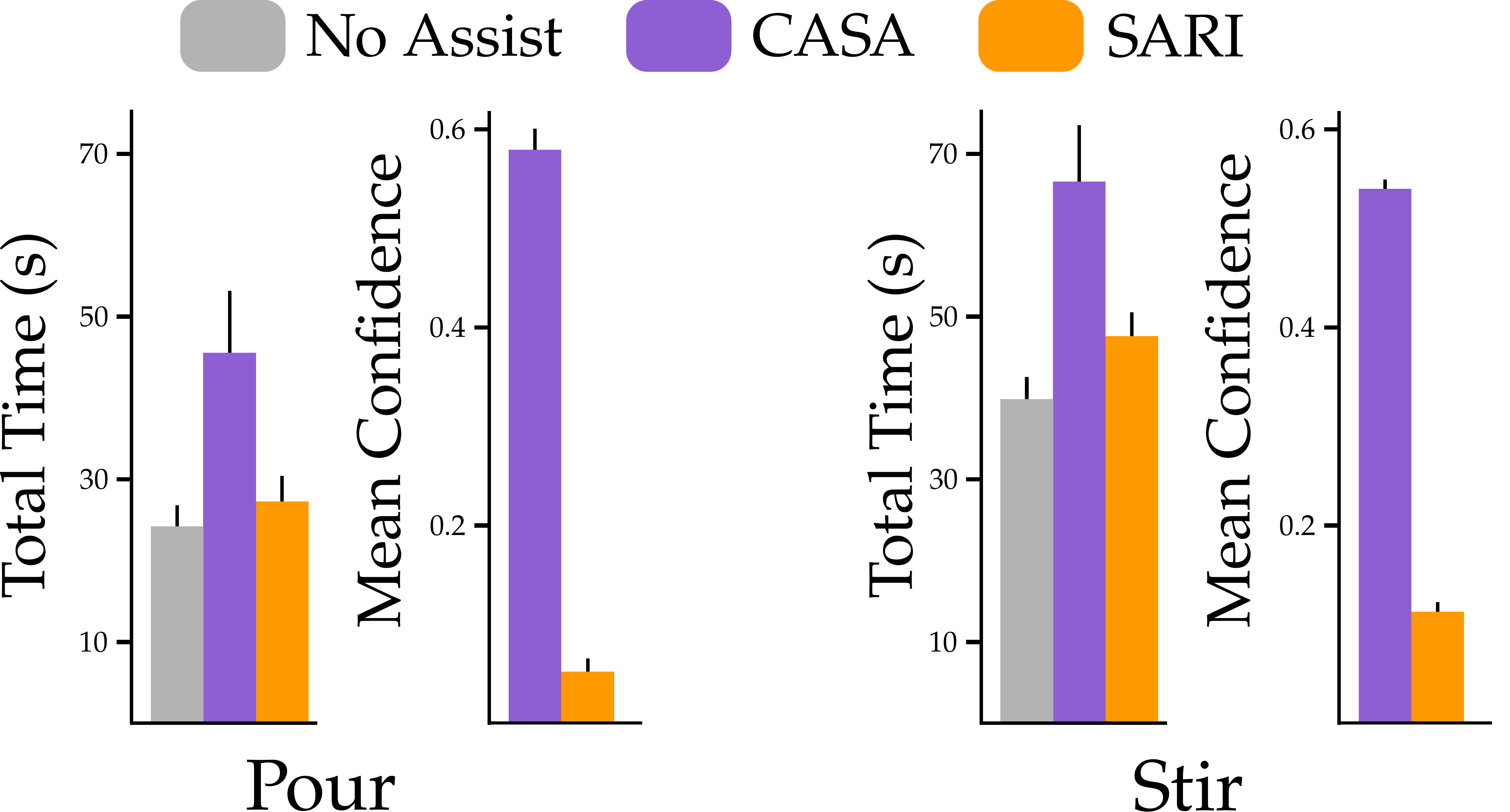}

		\caption{Objective results for the \textbf{unknown} tasks in our second user study. \textit{Mean Confidence} is the robot's average confidence $\beta$. Since these tasks are new to the robot, the robot should not be confident: ideally both \textit{Total Time} and \textit{Mean Confidence} will be minimized. \textbf{No Assist} always has zero confidence since the human is always in control of this robot. We observe that \textbf{CASA} significantly increased the \textit{Total Time} required by the human to complete these tasks (Pour: $F(2,87) = 16.6, p<.001$; Stir: $F(2,87) = 25.437, p< .001$). This increase in \textit{Total Time} can be attributed to the high \textit{Mean Confidence} \textbf{CASA} maintains (i.e., \textbf{CASA} incorrectly thinks it knows the task). By contrast, the \textbf{SARI} robot recognizes that it does not know the human's current task and correctly returns control so the human can act without interference.}
		\label{fig:user2_3}
	\end{center}
	\vspace{-1.1em}
\end{figure*}

\begin{figure}[t]
	\begin{center}
		\includegraphics[width=0.75\columnwidth]{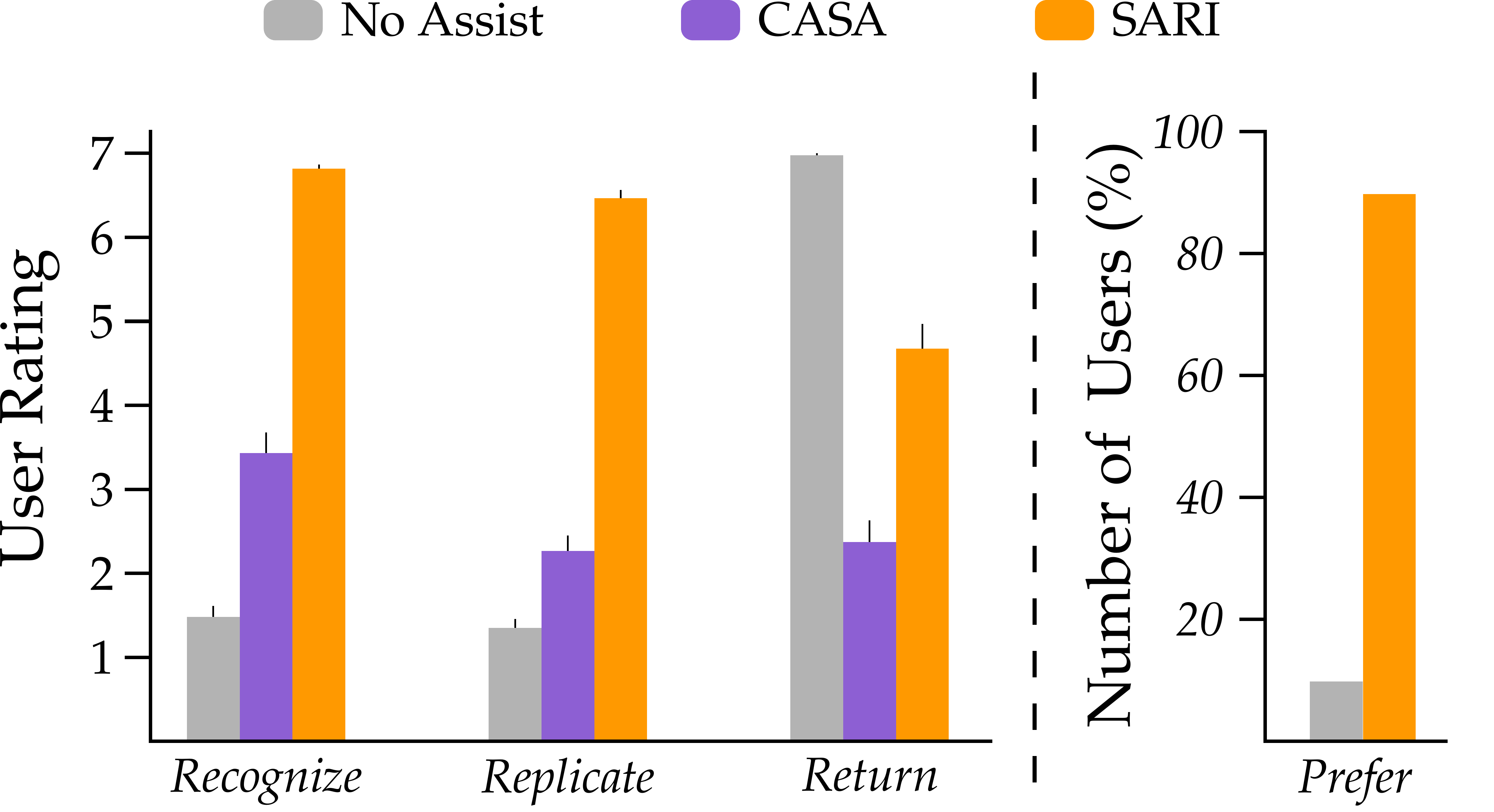}

		\caption{Subjective results for our second in-person user study. Similar to \fig{likert}, higher ratings indicate user agreement. Overall, participants found that \textbf{SARI} was able to better recognize their task and offer meaningful assistance. Additionally, a majority of users preferred using \textbf{SARI} over both \textbf{CASA} and \textbf{No Assist}. Scores for \textit{Recognize}, \textit{Replicate}, and \textit{Return} were obtained after participants completed each task with each method. For \textit{Prefer} we asked participants to select their favorite algorithm at the end of the experiment: $9$ out of $10$ users chose \textbf{SARI}. All differences shown here are statistically significant $(p < .01)$}
		\label{fig:likert2}
	\end{center}
	\vspace{-1em}

\end{figure}

In the previous user study we focused on the robot's ability to learn discrete goals and continuous skills. In our second user study we now focus on the effectiveness of the assistance offered by the robot on known tasks and its ability to \textit{return} control on new tasks. Figures~\ref{fig:user2_1}, \ref{fig:user2_2}, \ref{fig:user2_3}, and \ref{fig:likert2} correspond to this study. We compare our proposed approach with a previously seen baseline (\textbf{No Assist}) and a new state-of-the-art shared autonomy algorithm (\textbf{CASA}) \cite{zurek2021situational}. Similar to our method, \textbf{CASA} learns to assist users using previous demonstrations.

\p{Experimental Setup} Each participant teleoperated a Universal Robots UR$10$ robot arm to complete three continuous skills (see \fig{user2_1}): picking up a lemon and dropping it in a pan (\textbf{Drop}), picking up a can of soup and pouring it in a pan (\textbf{Pour}), and picking up a spatula and stirring the pan using a circular motion (\textbf{Stir)}. We emphasize that these tasks are continuous skills that cannot be easily reduced to a series of goals. To ensure that the human's actions can always steer the robot, we limit the robot's assistive action by restricting the robot's confidence ($\beta_{max} = 0.6$). 

The experimental procedure trades-off between teaching new tasks and then repeating known tasks. First, users completed \textbf{Drop} with a robot that was pre-trained only on the \textbf{Drop} task\footnote{In the interest of time, all models for SARI and CASA were trained using expert demonstrations. SARI required $<10$ minutes per model to be fully trained while CASA required over $8$ hours per model.}. Then --- using the same assistive models --- users completed \textbf{Pour}: during this interaction the robot knew \textbf{Drop} while \textbf{Pour} was an unknown task. Next, users completed \textbf{Pour} with a robot that was trained on demonstrations for both \textbf{Drop} and \textbf{Pour}. Following the same pattern, users completed the new \textbf{Stir} task. Here the known tasks are \textbf{Drop} and \textbf{Pour} and the unknown task is \textbf{Stir}. Finally users completed \textbf{Stir} with a model that was trained on all three tasks. Users attempted each known and unknown task with every method \textit{three} times. Overall, each individual user provided a total of $27$ demonstrations for the three known tasks and $18$ demonstrations for the two unknown tasks. 

\p{Independent Variables} Participants completed each known and unknown task three times with each method. We then compared the robot's assistance on known tasks and its ability to return control on unknown tasks across three different methods: \textbf{No Assist}, \textbf{CASA}, and \textbf{SARI}. \textbf{No Assist} did not offer any type of assistance to the users, while \textbf{CASA} and \textbf{SARI} learned to offer assistance. More specifically, \textbf{CASA} learns cost functions and policies for each new task using Guided Cost Learning \cite{finn2016guided}, and then infers which (if any) of the tasks the human is attempting to complete \cite{zurek2021situational}. Unlike \textbf{CASA}, our method (\textbf{SARI}) directly matches the human's past behaviors without inferring a cost function or learning a policy to minimize that cost.

\p{Dependent Measures -- Objective} For known tasks we measured \textit{Operating Time} and \textit{Opposing Time}, and for unknown tasks we plot \textit{Total Time} and \textit{Mean Confidence}. Here \textit{Operating Time} is the fraction of total time that the human provides input to the robot and \textit{Opposing Time} is the fraction of total time where the dot product between the human's action and the robot's action is negative (i.e., the fraction of time where the robot's assistance is not aligned with the human's commands). \textit{Total Time} is the time required to complete a task in seconds, and \textit{Mean Confidence} measures the robot's average confidence $\beta$ while the human is performing a task. When the user is performing a \textit{known} task, lower \textit{Operating Time} and \textit{Opposing Time} indicate that the robot is offering meaningful assistance to the user. In the case of \textit{unknown} tasks, lower \textit{Total Time} and \textit{Mean Confidence} indicate that the robot realizes its uncertainly and correctly returns control to the user.

\p{Dependent Measures -- Subjective} Similar to the study in Section~\ref{sec:user1} we administered a 7-point Likert scale survey (see Figure \ref{fig:likert2}). Questions were organized along four scales: how confident users were that the robot \textit{Recognized} their objective, how helpful the robot's behavior was (\textit{Replicate}), how trustworthy users thought the robot was (\textit{Return}), and if they would collaborate with the robot again (\textit{Prefer}). Users answered questions related to \textit{Recognize}, \textit{Replicate}, and \textit{Return} after completing each task with each method. At the end of the experiment (after trying each approach) users selected which of the three algorithms they \textit{Preferred}.

\p{Participants and Procedure}
We recruited $10$ non-disabled members from the Virginia Tech community to participate in our study ($5$ female, average age $27 \pm 4$ years). All participants provided informed written consent prior to the experiment under Virginia Tech IRB $\#20$-$755$. We used a within-subjects design and counterbalanced the order of the learning algorithms between subjects. Before the user study started each participant was given $10-15$ minutes to teleoperate the robot and familiarize themselves with the controls. The participants in this user study did not take part in the previous experiment from Section~\ref{sec:user1}.

\p{Hypotheses}
We tested two main hypotheses:
\begin{itemize}
    \item[] \textbf{H4.} \textit{For previously seen tasks, SARI will better assist than CASA or No Assist}
    \item[] \textbf{H5.} \textit{For new tasks, SARI will better return control to humans than CASA}
\end{itemize}

\p{Results -- Objective} The objective results from our second user study are shown in Figures~\ref{fig:user2_2}, \ref{fig:user2_3} and \ref{fig:likert2}. We separate our findings into two categories: performance when the robot has seen the task before (\textit{known}) and performance when the task is new (\textit{unknown}).

\smallskip \noindent \textit{Known tasks}: Results for the known tasks are shown in \fig{user2_2}. We performed a one way ANOVA to analyze these results. Post hoc comparisons revealed that across all three tasks \textbf{SARI} reduced the \textit{Operating Time} significantly more than \textbf{No Assist} or \textbf{CASA} (Drop: $F(2, 87) = 311.0, p < .001$; Pour: $F(2,87) = 278.7, p < .001$; Stir: $F(2,87) = 95.3, p < .001$). We also observed that the \textit{Opposing Time} for \textbf{SARI} across all three tasks was significantly lower than \textbf{CASA} (Drop: $F(2,87) = 368.6, p<.001$; Pour: $F(2,87) = 347.4, p<.001$; Stir: $F(2,87) = 349.6, p<.001$). Note that \textbf{No Assist} never opposes since the human is always in control. 

\smallskip \noindent \textit{Unknown tasks}: Results for the unknown tasks are shown in \fig{user2_3}. Remember that participants first performed \textbf{Pour} while the robot had only seen \textbf{Drop}, and then they first performed \textbf{Stir} while the robot was trained on \textbf{Drop} and \textbf{Pour}. Hence, here we do not expect the robot to assist the human: instead, the robot should recognize that the human is completing a new task and return control to the operator. We performed a one way ANOVA to analyze our results and determine whether the robot correctly returned control. We found that \textbf{CASA} increased the \textit{Total Time} required to complete both tasks (Pour: $F(2,87) = 16.6, p<.001$; Stir: $F(2,87) = 25.437, p< .001$). \textbf{CASA} increased the total time because this robot mistakenly thought the human was attempting to perform a known task rather than the unknown task. See the \textit{Mean Confidence}: the maximum allowed mean confidence was $0.6$, and \textbf{CASA} is close to this maximum confidence for both tasks (i.e., the \textbf{CASA} robot was almost entirely convinced it was performing a known task). Using paired $t$-tests we found that the mean confidence was significantly lower for \textbf{SARI} as compared to \textbf{CASA} (Pour: $t(29) = 40.928, p < .001$, Stir: $t(29) = 40.928, p < .001$).

\p{Results -- Subjective} We display the user's subjective responses to each algorithm in \fig{likert2}. After confirming that the scales were reliable (Cronbach's $\alpha > 0.7$), we grouped each scale into a single combined score and performed a one-way ANOVA on the results. We observe that across all three tasks users reported that \textbf{SARI} best \textit{recognized} their task ($F(2, 177) = 275.6, p < .001$) and \textit{replicated} their demonstrations to offer the most meaningful assistance ($F(2, 177) = 406.8, p < .001$). When it came to \textit{returning} control, \textbf{No Assist} was the gold standard; remember that for this method the human was always in control. However, here \textbf{SARI} was rated as significantly better than \textbf{CASA} ($F(2, 117) = 102.7, p < .001$). Overall, $9$ of the $10$ participants indicated that \textbf{SARI} was their preferred method for sharing autonomy with the robot.

\subsection{Assisting Users with Disabilities} \label{sec:user3}

\begin{figure}[t]
	\begin{center}
		\includegraphics[width=1.0\columnwidth]{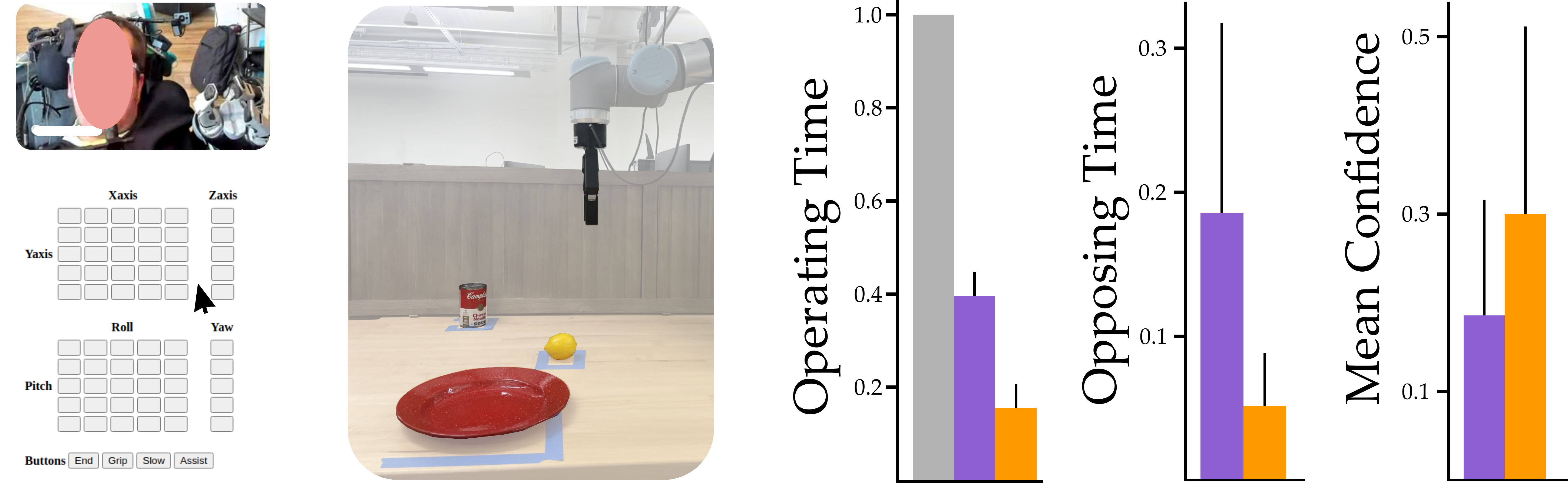}

		\caption{Experimental setup and objective results for our remote user study with one disabled participant. (Left) The participant used a web-based GUI to teleoperate the robot. The GUI was designed to mimic commercial interfaces \cite{kinova} and enabled the user to control the robot's end-effector velocity. (Center) We show one of the two camera angles that were streamed in real-time to the remote participant. (Right) Over the course of a one hour session the user completed \textbf{Drop} and \textbf{Pour} tasks (see \fig{user2_1}). Across both tasks \textbf{SARI} reduced the \textit{Operating Time} by maintaining a high \textit{Mean Confidence} and low \textit{Opposing Time}. These results suggest that \textbf{SARI} can assist adults who regularly use wheelchair-mounted robot arms.}
		\label{fig:user3_1}
	\end{center}

\end{figure}

In our third and final pilot study we explore how our approach assists a disabled user who operates robot arms on a daily basis. Similar to Section~\ref{sec:user2}, the participant interacted with a robot arm that either (a) did not provide any assistance, (b) assisted using the state-of-the-art CASA approach \cite{zurek2021situational}, or (c) assisted using our proposed SARI algorithm. This pilot study was conducted remotely: the participant observed the robot in real-time through a live video feed, and remotely teleoperated the robot using a web-based GUI (see \fig{user3_1}).

\p{Experimental Setup and Independent Variables} The participant teleoperated the Universal Robots UR$10$ robot arm to perform two tasks from Section~\ref{sec:user2}: \textbf{Drop} and \textbf{Pour}. The robot had expert demonstrations for both tasks (i.e., both tasks were \textit{known} by the robot). Over a one hour session we collected a total of $6$ demonstrations across all tasks and methods from the participant. We compared the effectiveness of the robot's assistance when using \textbf{No Assist}, \textbf{CASA}, or \textbf{SARI}. 

\p{Dependent Measures} We measured \textit{Operating Time}, \textit{Opposing Time}, and \textit{Mean Confidence}. Recall that \textit{Operating Time} is the the fraction of total time where the human is providing inputs to the robot, \textit{Opposing Time} is the fraction of total time where the human and robot actions are in opposite directions, and \textit{Mean Confidence} is the robot's average confidence that it should provide assistance ($\beta$). Since the robot is performing known tasks, an ideal system will maintain high \textit{Mean Confidence}, partially automate the motion, and reduce \textit{Operating Time} and \textit{Opposing Time}.

\p{Participants and Procedure} We recruited one remote participant for this pilot study. The participant provided informed written consent under Virginia Tech $\#20$-$755$. Since this user study was conducted online, the participant used a web-based GUI to control the robot (see \fig{user3_1}). This GUI includes a table of buttons to control the position and orientation of the end-effector: we designed the GUI to mimic online interfaces for commercial wheelchair-mounted robot arms \cite{kinova}. The participant moved their cursor and pressed buttons on the GUI using a joystick that was integrated with their wheelchair. For the first $15$ minutes of the experiment the participant familiarized themselves with our robot, environment, and controls. During this time we also worked with the participant to find the best camera placements.

\p{Results} \fig{user3_1} presents our results. Remember that the robot is performing a task that it has seen before; ideally the assistive robot will realize this is a known task and help the user. We observed that --- similar to the previous user study --- both \textbf{CASA} and \textbf{SARI} reduced the \textit{Operating Time}. However, we also noticed that across both tasks \textbf{SARI} provided better assistance by maintaining high \textit{Mean Confidence} and keeping the \textit{Opposing Time} to a minimum. While we only had one participant in this pilot study, the results obtained suggest that \textbf{SARI} can offer meaningful assistance to our target population without requiring pre-defined tasks.

%% file: conclusion.tex
\section{Conclusion}

State-of-the-art shared autonomy algorithms often rely on prior knowledge: e.g., the robot needs to know all of the human's potential tasks \textit{a priori}, or the robot is constrained to actions that assist for previously seen tasks. In this paper we introduce an alternate framework for shared autonomy that leverages the \textit{repeated} nature of everyday human-robot interaction. Our insight is that --- if an assistive arm is teleoperated through similar tasks many times --- the robot should learn to \textit{partially automate} those tasks. Our approach (SARI) contains separate models that (a) learn to recognize the human's current task, (b) replicate the human's behavior from past task-related interactions, and (c) return control back to the human when the robot is unsure. We leveraged stability analysis that combines learning with control to demonstrate that the error between the human's goal and the SARI robot is uniformly ultimately bounded. We then conducted simulations to support our theoretical error bounds, compare our approach to interactive imitation learning baselines, and explore the capacity of SARI to learn new tasks. Finally, we performed three user studies to demonstrate that SARI assists for both discrete goals and continuous skills, shows greater ability to recognize tasks and return control, and provides meaningful assistance to users with disabilities. Overall, our theoretical and experimental analysis suggests that SARI \textit{personalizes} to the current user, and can learn to share autonomy for the tasks \textit{that user} often performs.

\p{Limitations} So far we have focused on how assistive robot arms can adapt to their human users. But as the robot arm gets better at sharing autonomy, the human will also \textit{co-adapt} and modify their own teleoperation strategy. For example, once the human is confident the robot recognizes their current task, the user may stop providing joystick inputs and rely on the robot entirely. One approach to circumvent this issue is by only storing interactions if the human provides corrective actions. However, this co-adaptation is not explicitly accounted for in our approach.

Another potential limitation of SARI may be its capacity. From Section~\ref{sec:sim_capacity} we recognize that the robot's assistance decreases as the number of skills increases. More generally, it may not be feasible for a single model to learn assistance for all of the human's everyday tasks. One practical solution is switching models depending on \textit{context}. For example, we could train one instance of SARI to assist for cooking tasks, and another instance of SARI to assist for dining tasks. During run-time the robot selects which SARI models to use based on the current context (e.g., cooking or eating).

%% file: appendix.tex
\section{Appendix}

\subsection{Details for Error Bounds from Section~\ref{sec:theory}}
Below we provide additional details for the error bounds that were presented in Section~\ref{sec:theory}.

\p{Derivation of Equation \ref{eq:T5}} From \eq{T1} we know that the arbitration constant $\beta (s, a_{\mathcal{H}})$ is defined as: 

\begin{equation*}
        \beta(s, a_{\mathcal{H}})
    = \frac{1}{\sqrt{2\pi\sigma_{\mathcal{D}}^2}}\exp{ \left( \frac{-\big(a_{\mathcal{H}}-(g-s)\big)^2}{2\sigma_{\mathcal{D}}^2}\right)}
\end{equation*}

Taking into consideration that ${a_{\mathcal{H}} \sim \mathcal{N}\big((g^*-s), \sigma_{\mathcal{H}}^2)\big)}$ and recalling that the law of the unconscious statistician (LOTUS) \cite{schervish2014probability} states that for a function $g(X)$ of a random variable $X$, the expected value $\mathbb{E}[g(X)] = \int_{-\infty}^{\infty}g(x)f_X(x) \, dx$, where $f_X(x)$ is the probability density function of $X$. We take the expectation of \eq{T1} and substitute the above to obtain:

\begin{equation}
    \mathbb{E}[\mathcal{\beta}] = \int_{-\infty}^{\infty} \frac{1}{\sqrt{2\pi \sigma^2_{\mathcal{D}}}} \exp{\bigg(-\frac{1}{2}\frac{\big(a_{\mathcal{H}} - (g-s)\big)^2}{\sigma_{\mathcal{D}}^2}\bigg)} \cdot \frac{1}{\sqrt{2\pi \sigma^2_{\mathcal{H}}}}\exp{\bigg(-\frac{1}{2}\frac{\big(a_{\mathcal{H}} - (g^*-s)\big)^2}{\sigma_{\mathcal{H}}^2}\bigg)} \, da_{\mathcal{H}}\\
\end{equation}

Upon further simplification we obtain:

\begin{equation*}
\mathbb{E}[\beta] =
  \dfrac{1}{\sqrt{2\pi(\sigma_{\mathcal{D}}^2 + \sigma_{\mathcal{H}}^2)}}\exp{ \left( \dfrac{-(g^*-g)^2}{2(\sigma_{\mathcal{D}}^2 + \sigma_{\mathcal{H}}^2)}\right)}
\end{equation*}

\p{Derivation of Equation \ref{eq:T41}} When taking the expectation of $\mathbb{E}[\beta \alpha_{\mathcal{H}}]$, we utilize LOTUS to obtain:

\begin{equation*}
    \mathbb{E}[\beta a_{\mathcal{H}}] = \int_{-\infty}^{\infty} \frac{a_\mathcal{H}}{\sqrt{2\pi \sigma^2_{\mathcal{D}}}}\exp{\bigg(-\frac{1}{2}\frac{\big(a_\mathcal{H} - (g-s)\big)^2}{\sigma_\mathcal{D}^2}\bigg)} \cdot \frac{1}{\sqrt{2\pi \sigma^2_\mathcal{H}}}\exp{\bigg(-\frac{1}{2}\frac{\big(a_\mathcal{H} - (g^*-s)\big)^2}{\sigma_\mathcal{H}^2}\bigg)} \, da_\mathcal{H}
\end{equation*}

Upon simplification we obtain:

\begin{equation}\label{eq:A12}
    \mathbb{E}[\beta a_{\mathcal{H}}] =
    \frac{1}{2\pi \sigma_\mathcal{D} \cdot \sigma_\mathcal{H}} \int_{-\infty}^{\infty} a_\mathcal{H} \cdot\exp{\bigg(-\frac{1}{2}\frac{\big(a_\mathcal{H} - (g-s)\big)^2}{\sigma_\mathcal{D}^2} -\frac{1}{2}\frac{\big(a_\mathcal{H} - (g^*-s)\big)^2}{\sigma_\mathcal{H}^2}\bigg)} \, da_\mathcal{H}
\end{equation}

By solving for the integral in \eq{A12} we obtain \eq{T41}.

\p{Derivation of Equation \ref{eq:T10}} Recall that: 

\begin{equation}\label{eq:A13}
    \beta(\bm{s}, \bm{a_\mathcal{H}}) = \frac{1}{(2\pi)^{d/2} |\bm{\Sigma_\mathcal{D}}|^{1/2}} \exp{\left(-\frac{1}{2} \cdot (\bm{a_\mathcal{H}} - (\bm{\bm{g^*} - s}))^T \bm{\Sigma_\mathcal{D}}^{-1} (\bm{a_\mathcal{H}} - (\bm{\bm{g^*} - s}))\right)}
\end{equation}

We rewrite \eq{A13} using the canonical parameterization as:

\begin{equation*}
    \beta(\bm{s}, \bm{a_\mathcal{H}}) =  \exp{\left(\xi_\mathcal{D} + \bm{\eta_\mathcal{D}}^T \cdot \bm{a_\mathcal{H}} - \frac{1}{2} \bm{a_\mathcal{H}}^T\bm{\Lambda_\mathcal{D}} \bm{a_\mathcal{H}}\right)}
\end{equation*}

where $\bm{\Lambda_\mathcal{D}} = \bm{\Sigma_\mathcal{D}}^{-1}$, $\bm{\eta_\mathcal{D}} = \bm{\Sigma_\mathcal{D}}^{-1}(\bm{g^* - s}$), and $\xi_\mathcal{D} = -\frac{1}{2}(d\log2\pi - \log|\bm{\Lambda}_\mathcal{D}| + \bm{\eta}^T_{\mathcal{D}}\bm{\Lambda_\mathcal{D}}^{-1}\bm{\eta}_{\mathcal{D}})$. To compute $\mathbb{E}(\beta)$, we utilize LOTUS:

\begin{equation*}
    \mathbb{E}[\beta] = \int_{-\infty}^{\infty} \exp{\left(\xi_\mathcal{D} + \bm{\eta_\mathcal{D}}^T \cdot \bm{a_\mathcal{H}} - \frac{1}{2} \bm{a_\mathcal{H}}^T\bm{\Lambda_\mathcal{D}} \bm{a_\mathcal{H}}\right)} \cdot \exp{\left(\xi_\mathcal{H} + \bm{\eta_\mathcal{H}}^T \cdot \bm{a_\mathcal{H}} - \frac{1}{2} \bm{a_\mathcal{H}}^T\bm{\Lambda_\mathcal{H}} \bm{a_\mathcal{H}}\right)}
\end{equation*}    

Upon further simplification we obtain:
\begin{equation*}
    \mathbb{E}[\beta]
    = \int_{-\infty}^{\infty} \exp{\left(\xi_D + \xi_H - \xi + \xi + \bm{\eta}^T\bm{a_h} - \frac{1}{2} \bm{a_h}^T\bm{\Lambda} \bm{a_h} \right)}
\end{equation*}

where $\bm{\Lambda} = \bm{\Lambda_D} + \bm{\Lambda_H}$, $\bm{\eta} = \bm{\eta_D} + \bm{\eta_H}$, and $\xi = -\frac{1}{2}(d\log2\pi - \log|\bm{\Lambda}| + \bm{\eta}^T\bm{\Lambda}^{-1}\bm{\eta})$. With substitution and simplification we obtain:

\begin{equation}\label{eq:A14}
    \mathbb{E}[\beta]
    = \frac{1}{(2\pi)^{d/2} |\bm{\Sigma}_\mathcal{D}+\bm{\Sigma}_\mathcal{H}|^{1/2}}\exp{\left(-\frac{1}{2}(\bm{g} - \bm{g}^*)^T (\bm{\Sigma}_\mathcal{D} + \bm{\Sigma}_\mathcal{H})^{-1}(\bm{g} - \bm{g}^*)\right)}
\end{equation}

\eq{A14} can be simplified and rewritten as \eq{T10}.

\p{Proof for Theorem 4} Because we have assumed that $\mathbb{E}[\beta] < \beta_{max}$, we set $\beta = \beta(s, \boldsymbol{a_{\mathcal{H}}})$. Note that $\beta$ depends upon $\boldsymbol{a_{\mathcal{H}}}$. We transform $\beta(s, \boldsymbol{a_{\mathcal{H}}})$ using the canonical parameterization and apply LOTUS to compute the expectation as:

\begin{equation}
    \mathbb{E}[\beta \bm{a_\mathcal{H}}] = \int_{-\infty}^{\infty} \bm{a_\mathcal{H}} \exp{\left(\xi_\mathcal{D} + \bm{\eta_\mathcal{D}}^T \cdot \bm{a_\mathcal{H}} - \frac{1}{2} \bm{a_\mathcal{H}}^T\bm{\Lambda_\mathcal{D}} \bm{a_\mathcal{H}}\right)} \cdot \exp{\left(\xi_\mathcal{H} + \bm{\eta_\mathcal{H}}^T \cdot \bm{a_\mathcal{H}} - \frac{1}{2} \bm{a_\mathcal{H}}^T\bm{\Lambda_\mathcal{H}} \bm{a_\mathcal{H}}\right)}
\end{equation}

Upon further simplification and reparameterization we obtain:
\begin{equation}
    \mathbb{E}[\beta \bm{a_\mathcal{H}}]
    = \frac{(\bm{\Sigma}_H + \bm{\Sigma}_D)^{-1}(\bm{\Sigma}_D(\bm{g}^*-\bm{s}) + \bm{\Sigma}_H (\bm{g}-\bm{s}))}{(2\pi)^{d/2} |\bm{\Sigma}_D+\bm{\Sigma}_H|^{1/2}}\exp{\left(-\frac{1}{2}(\bm{g} - \bm{g}^*)^T (\bm{\Sigma}_D + \bm{\Sigma}_H)^{-1}(\bm{g} - \bm{g}^*)\right)}
\end{equation}

This can be rewritten as:
\begin{equation} \label{eq:A1}
\mathbb{E}[\beta \boldsymbol{a_{\mathcal{H}}}] = \frac{\Sigma^{-1}\big(\Sigma_{\mathcal{D}}(\boldsymbol{g^*} - \boldsymbol{s})+\Sigma_{\mathcal{H}}(\boldsymbol{g} - \boldsymbol{s})\big)}{\sqrt{(2\pi)^d \det\Sigma}} \cdot \exp{\left(-\frac{1}{2}\|\boldsymbol{g^*} - \boldsymbol{g}\|^2_{\Sigma^{-1}}\right)}
\end{equation}
where $\Sigma = \Sigma_{\mathcal{D}} + \Sigma_{\mathcal{H}}$.
Substituting \eq{T10} and \eq{A1} back into \eq{T9}, we have $\mathbb{E}[\dot{V}(t)] < 0$ when:
\begin{equation} \label{eq:A2}
    \|\boldsymbol{g^*}-\boldsymbol{s}\|^2 > \mathbb{E}[\beta] \cdot 
    (\boldsymbol{g^*} - \boldsymbol{s})^T \Sigma^{-1}\Sigma_{\mathcal{D}}(\boldsymbol{g^*} - \boldsymbol{g})
\end{equation}
Here we apply the Cauchy–Schwarz inequality to obtain:
\begin{equation} \label{eq:A21}
    \|\boldsymbol{g^*}-\boldsymbol{s}\|^2 > \mathbb{E}[\beta] \|\boldsymbol{g^*} - \boldsymbol{s}\| \cdot  \|\Sigma^{-1}\Sigma_{\mathcal{D}}(\boldsymbol{g^*} - \boldsymbol{g})\|
\end{equation}
From the spectral theorem we know that $\|Ax\| \leq \lambda_{max}(A)\|x\|$, where $\lambda_{max}(A)$ is the largest eigenvalue of the positive semidefinite matrix $A$. We substitute this inequality in \eq{A1} to obtain a more relaxed constraint. Specifically, we find that $\mathbb{E}[\dot{V}(t)] < 0$ if the following inequality holds:
\begin{equation} \label{eq:A3}
    \|\boldsymbol{g^*}-\boldsymbol{s}\|^2 > \|\boldsymbol{g^*} - \boldsymbol{s}\| \cdot \lambda \mathbb{E}[\beta] \cdot \|\boldsymbol{g^*} - \boldsymbol{g}\|
\end{equation}
where $\lambda$ is the maximum eigenvalue of $\Sigma^{-1}\Sigma_D$. Rearranging this result yields \eq{T12}. We conclude that $\mathbb{E}[\dot{V}(t)] < 0$ when \eq{T12} is satisfied, and it therefore follows that the human-robot system is uniformly ultimately bounded.

\p{Ultimate Bounds} In our proofs for Theorems~1--4 we have shown that the SARI system is uniformly ultimately bounded, and we have listed the ultimate bounds. However, we have not formally demonstrated \textit{why} Equations~(\ref{eq:T6}), (\ref{eq:T7}), (\ref{eq:T11}), and (\ref{eq:T12}) are the ultimate bounds. Here we provide a more rigorous derivation for these results. We focus on Theorem 4, but the same approach applies to each of our Theorems.

Recall from \eq{T8} that the Lyapunov candidate function depends on the error $\boldsymbol{e}$, and remember that $\boldsymbol{e}=\boldsymbol{g^*}-\boldsymbol{s}$. Let $\alpha_1$ and $\alpha_2$ be two class $\kappa$ functions such that:
\begin{equation} \label{eq:A4}
    \alpha_1\big(\|\boldsymbol{e}\|\big) \leq V(\boldsymbol{e}) \leq \alpha_2\big(\|\boldsymbol{e}\|\big)
\end{equation}
Here the ultimate bound on error $\boldsymbol{e}$ can be taken as \cite{khalil2002nonlinear}:
\begin{equation} \label{eq:A5}
    b = \alpha_1^{-1}\big(\alpha_2(\mu)\big)
\end{equation}
where $\mu > 0$ is selected such that $\mathbb{E}[\dot{V}(t)] < 0$ for all $\|\boldsymbol{e} \| > \mu$. Looking back at \eq{T12} and the previous proof from the Appendix, we have already identified $\mu = \lambda \mathbb{E}[\beta] \cdot \|\boldsymbol{g^*} - \boldsymbol{g}\|$. We now propose $\alpha_1 = \alpha_2 = \frac{1}{2}\|\boldsymbol{e}\|^2$. These choices are valid because (a) they satisfy \eq{A4} and (b) they are class $\kappa$ functions. Plugging $\mu$, $\alpha_1$, and $\alpha_2$ back into \eq{A5}, the ultimate bound is:
\begin{equation}
    b = \lambda \mathbb{E}[\beta] \cdot \|\boldsymbol{g^*} - \boldsymbol{g}\|
\end{equation}
Intuitively, this result means that the expected error between the human's new goal $\boldsymbol{g^*}$ and the robot's state $\boldsymbol{s}$ will eventually become smaller than $b$, and will remain smaller than $b$ for the rest of the interaction \cite{spong2006robot}.

\subsection{Implementation Details}

 Our code can be found here: \url{https://github.com/VT-Collab/repeated-shared-autonomy}.

 \p{Data collection} Across all three user studies we record the robot's state and the human's actions. Depending on the task, we collect the robot's joint positions, robot's Cartesian position, state of the gripper, current mode of teleoperation (rotation or translation), and current mode of operation (fast or slow mode). In-person users utilized the joysticks on a Logitech Gamepad F310 controller to provide their inputs. Since the controller is equipped with only two joysticks, users can toggle between translational and rotational inputs using one of the buttons on the controller. Remote users used a web-based GUI that had individual buttons that could be used to command translational and rotational velocities. This GUI also had additional buttons to open and close the gripper, and to speed up or slow down the robot. The inputs from the joystick and the web GUI are converted into a $6$-D vector of velocity inputs and stored as the human's current action. 

 \p{Data augmentation} We augment the data we receive from the demonstrations using Gaussian noise. We create $5$ additional samples for each sample we collect by injecting Gaussian noise with zero mean and a small variance. We use this augmented data to train our method as well as all the baselines that require human demonstrations for training. 
 
\p{Deformations for the discriminator} To create samples that represent unseen behavior for our discriminator, we randomly apply a small force to the demonstrations we collected from the human. This force alters the start, end and the shape of the initial trajectory. We use previous work by \cite{losey2017trajectory} to generate these deformations. Additional details on our implementation can be found in our code.
 
 \p{Network architecture} We use fully connected networks for the encoder, decoder and the discriminator in our method. While we varied the number of hidden layers and the number of neurons in each layer throughout the project, we found the best results when our encoder consisted of $5$ hidden layers, the decoder consisted of $4$ hidden layers, and the discriminator consisted of $4$ hidden layers. Additional information on our specific implementations can be found in the code repository. 

\p{Computational Requirements} Throughout the project we use various versions of Pytorch and Python in our environment. Most of our experiments were run on an Intel PC with an i7-8559U processor and 32GB of RAM. We store the human interaction data as Python pickle files and each interaction requires 72kB of memory. Our Pytorch models were also saved as pickle files and required 42.7kB of memory. No GPU was used for our experiments.

 \p{Baselines} For DAgger-based baselines we use fully connected networks similar to our method. For DropoutDagger we use dropout probability of $0.1$ and for EnsembleDagger we train and use an ensemble of $20$ models. To obtain the network's confidence ($\beta$) we scale the variance within the model's action for DropoutDagger and variance between models for EnsembleDagger with a constant value. This constant is obtained by tuning the robot's performance while performing a learned task. For CASA we use a combination of Guided Cost Learning (GCL) and Soft-Actor Critic (SAC) \cite{haarnoja2018soft}. We implement SAC in-house and modify a pre-existing repository for training using GCL. The original repository can be found here: \url{https://github.com/NinaWie/guided-cost-learning}, and our specific implementation can be found in our code repository.